\documentclass[10pt,twocolumn,letterpaper]{article}

\usepackage{iccv}              

\usepackage{multirow}
\usepackage{makecell}

\usepackage[accsupp]{axessibility}

\usepackage{algorithm}
\usepackage{algpseudocode}
\usepackage{mathtools}

\definecolor{iccvblue}{rgb}{0.21,0.49,0.74}
\usepackage[pagebackref,breaklinks,colorlinks,allcolors=iccvblue]{hyperref}

\algnewcommand\algorithmicforeach{\textbf{for all}}
\algdef{S}[FOR]{ForEach}[1]{\algorithmicforeach\ #1\ \algorithmicdo}

\def\paperID{6605} 
\def\confName{ICCV}
\def\confYear{2025}

\newcommand{\ourname}{SpinMeRound}

\newcommand{\allaround}{$360^{\circ}$}
\DeclareMathOperator*{\argmax}{arg\,max}  

\title{\ourname{}: Consistent Multi-View Identity Generation Using Diffusion Models}

\author{Stathis Galanakis$^{1}$ \hspace{1.5cm}
\and
Alexandros Lattas$^{1}$ \hspace{1.5cm}
\and
Stylianos Moschoglou$^{1}$ 
\and
\hspace{1.5cm} Bernhard Kainz$^{1,2}$ \hspace{1.5cm}
\and
Stefanos Zafeiriou$^{1}$
\and
\\
$^{1}$Imperial College London, UK \\
$^{2}$FAU Erlangen–Nürnberg, Germany
\vspace{-0.5cm}
}

\begin{document}
\twocolumn[{
\renewcommand\twocolumn[1][]{#1}%
\maketitle
\begin{center}
    \centering
    \captionsetup{type=figure}
    \includegraphics[width=\textwidth]{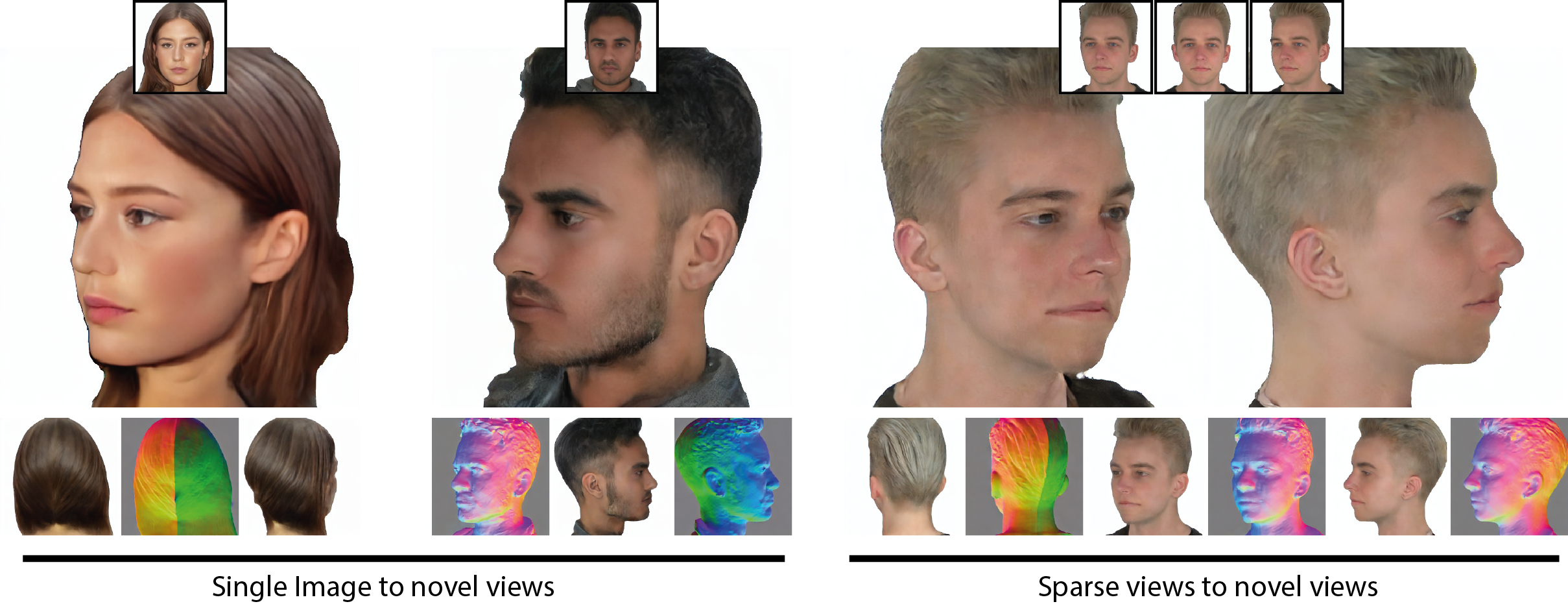}    
    \captionof{figure}{\ourname{} is a multiview diffusion model which generates human portraits from novel viewpoints.
    Given a single or multiple views, our method produces high-fidelity images along with  precise surface normals, ensuring accurate 3D consistency across perspectives.}

\end{center}}]
\begin{abstract}
Despite recent progress in diffusion models, generating realistic head portraits from novel viewpoints remains a significant challenge in computer vision.
Most current approaches are constrained to limited angular ranges, predominantly focusing on frontal or near-frontal views.
Moreover, although the recent emerging large-scale diffusion models have been proven robust in handling 3D scenes,
they underperform on facial data, given their complex structure and the uncanny valley pitfalls.
In this paper, we propose \ourname{}, a diffusion-based approach designed to generate consistent and accurate head portraits from novel viewpoints.
By leveraging a number of input views alongside an identity embedding, our method effectively synthesizes diverse viewpoints of a subject whilst robustly maintaining its unique identity features.
Through experimentation, we showcase our model's generation capabilities in full head synthesis, while beating current state-of-the-art multi-view diffusion models.
Project page: \href{https://spin-me-round.github.io/}{spin-me-round.github.io}.

\end{abstract}    
\section{Introduction}
\label{sec:intro}

Recent advancements in deep learning have brought significant progress to fundamental computer vision tasks, notably in image and video generation~\cite{Rombach_2022_CVPR, blattmann2023videoldm}. 
The introduction of Diffusion Models~\cite{ho2020denoising,NEURIPS2021_49ad23d1} has transformed these domains by enabling the generation of high-quality visual content, facilitated by the use of large-scale datasets~\cite{Rombach_2022_CVPR,blattmann2023stablevideodiffusionscaling}. 
Despite these developments, the task of generating consistent and accurate head portraits from a single input image still remains a challenging problem. 
This difficulty is primarily attributed to the limited availability of comprehensive 3D facial datasets, which constrains the training of models capable of reliably capturing and reconstructing the detailed structure and variations of human faces. 

A common practice for modeling human heads from a single image, incorporates the use of a 3D Morphable Model (3DMM)~\cite{Blanz1999AMM, booth20163d, booth2018large, FLAME:SiggraphAsia2017}, to represent the facial shape along with an appearance model~\cite{Gecer_2019_CVPR, DECA:Siggraph2021, lattas2023fitme}.
However, given typical training data and the difficulty in modeling complex hairstyles,
these methods focus solely on the facial region and avoid or miss the full head and hair.
The seminal work of Neural Radiance Fields (NeRF)~\cite{mildenhall2020nerf}
led to an explosion of works on neural rendering of scenes that could not easily be modeled with textured meshes.
In addition, pairing such implicit representations with generative models led to a wide variety of approaches~\cite{athar2022rignerf, Or-El_2022_CVPR, Chan_2022_CVPR} that pushed the boundaries on the synthesis of novel facial views, achieving high quality and control.
More recently, Panohead~\cite{an2023panohead} first showcased high-quality \allaround{} head portrait synthesis.
However, because of its adaptive camera training scheme, the back-head synthesis typically contains many artifacts~\cite{li2024spherehead},
and its inversion on ``in-the-wild'' images is challenging and requires complex fine-tuning ~\cite{roich2021pivotaltuninglatentbasedediting}.

Recently, diffusion models \cite{NEURIPS2021_49ad23d1} demonstrated superior performance over GAN-based methods in image generation and have achieved great quality in human generation tasks \cite{Rombach_2022_CVPR}.
However, achieving multi-view consistency remains a significant challenge. 
Despite the lack of accurate 3D datasets, recent advances, such as Score Distillation techniques~\cite{poole2022dreamfusion, zhou2024headstudio}, represent an initial step toward leveraging the 2D generation strengths of diffusion models to construct 3D content without extra training.
Nevertheless, these approaches are computationally intensive, require intricate constraints, and do not consistently yield photorealistic results.
Meanwhile, multi-view diffusion architectures, employing video diffusion models as a backbone framework~\cite{voleti2024sv3dnovelmultiviewsynthesis, melaskyriazi2024im3diterativemultiviewdiffusion}, still remain very resource-demanding for generating a single novel view, as they rely on comprehensive camera trajectories to generate coherent central objects.
Closer to our work, Zero123~\cite{liu2023zero1to3} introduced a view-conditioned diffusion model that incorporates view features and camera information into the diffusion process.
However, the generated images lack strong multi-view consistency and are of low quality, which restricts their performance in photorealistic 3D generation.
Although DiffPortrait3D~\cite{Gu_2024_CVPR} enables novel view generation through a diffusion process, it still focuses only on near-front views and utilizes a image-conditioned camera mechanism.
Other closely related works are Era3D~\cite{li2024era3d} and Morphable Diffusion~\cite{chen2024morphable}, which generate fixed camera viewpoints, thus limiting their ability to produce full head portraits.
The recently proposed Cat3D~\cite{gao2024cat3d} presents a promising solution by efficiently integrating a number of input views with specified camera poses to achieve a consistent novel viewpoint generation. 
It integrates 3D attention layers for efficiently sharing common information between all views, along with robust camera pose feature maps.
Still, Cat3D is limited by its lack of focus on human generation and is currently unavailable as an open-source tool.

In this paper, we present \ourname{}, a multi-view diffusion model designed to generate novel high-fidelity views of a given human face.
In addition to facial images, our model also generates the corresponding normals, 
which are typically available for human data and, as we show, improve the model's performance and consistency on intricate facial features.
Moreover, we show that conditioning the model on an identity embedding and one or more input views of a subject during inference,
we can not only sample but also reconstruct multi-view consistent images from ``in-the-wild'' facial images.
Our method can accurately synthesize photorealistic head portraits from various angles
while preserving essential identity characteristics, which can be used to represent or reconstruct 3D scenes.
Given the lack of open-source large-scale multi-view head datasets and the problematic nature of the permissions of such datasets, our method is solely trained on synthetic data acquired using Panohead~\cite{an2023panohead}, making this work accessible to experiment and build with.
Overall, in this paper:
\begin{itemize}
    \item We introduce \ourname{}, a multi-view diffusion model conditioned on identity embeddings and a number of views that generates novel perspectives of an input subject and their respective normals.
    \item We present a novel sampling strategy which, given a single ``in-the-wild'' facial image, generates consistent views encompassing the whole head.
    \item We explore its potential by comparing it with current state-of-the-art multi-view diffusion-based methods and showcase superior results in full-head portrait generation.
\end{itemize}

\section{Related Work}
\label{sec:relatedwork}
\subsection{Face Modeling}

Extensive research has been dedicated to representing 3D faces through a combination of texture maps and 3D meshes, starting with fundamental 3D Morphable Models (3DMM)~\cite{Blanz1999AMM,booth20163d, booth2018large, FLAME:SiggraphAsia2017}.
These approaches, however, primarily focus on accurately capturing only the frontal region of the head whilst lacking in integrating finer head details such as hair, wrinkles and wearable items.
Dealing with this, recent studies have integrated implicit representations for 3D face modeling. Methods such as ~\cite{zheng2022imface, giebenhain2023nphm, orel2022styleSDF} integrate
Signed Distance Functions (SDFs) whereas other~\cite{Chan_2022_CVPR, athar2022rignerf,  Galanakis_2023_WACV, yenamandra2021i3dmm} use Neural Radiance Field (NeRF) models~\cite{mildenhall2020nerf} for generating photorealistic results.
Levering the powerful triplane representation~\cite{Chan_2022_CVPR}, RTRF~\cite{trevithick2023} generates near-frontal views in real-time whilst Panohead~\cite{an2023panohead} introduces an adaptive camera training strategy for enabling full head portrait synthesis.
An extension of it is 3DPortraitGAN~\cite{wu20233dportraitgan} focusing on all-around upper body generation. 
As this work focuses only on full-head generation, Panohead can effectively be used to acquire synthetic portrait datasets.

\def \var {1}
\begin{figure*}[!t]
\begin{center}
 \begin{subfigure}[b]{\var\textwidth}
      \includegraphics[width=\textwidth]{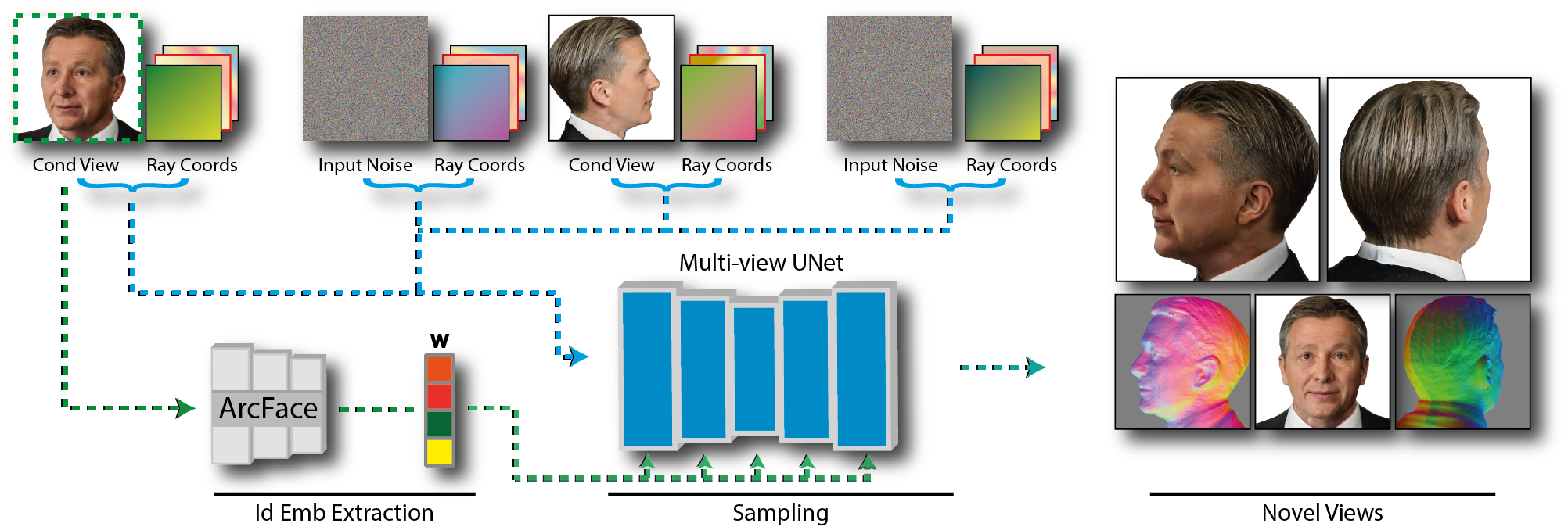}
 \end{subfigure}
  \caption{
    Overview of \ourname{}: Starting with a number of input conditioning views, the identity embedding $\mathbf{W}$ is extracted via a Face Recognition network (ArcFace~\cite{deng2019arcface}).
    Both the conditioning and target views are then encoded and combined with corresponding ray coordinate maps that represent camera poses.
    After the sampling step, our method synthesizes photorealistic images from novel angles, along with their associated shape normals $\mathcal{N}$.
    }
\label{fig:method}
\end{center}
\vspace{-0.7cm}
\end{figure*}

\subsection{Diffusion Models}
\paragraph{Face Generation}
Diffusion models (DMs)~\cite{ho2020denoising} have proven their generative abilities by beating the well-established GANs in image synthesis tasks~\cite{NEURIPS2021_49ad23d1}.
The availability of large-scale datasets has revolutionized text-to-image generation~\cite{Rombach_2022_CVPR} and video generation~\cite{blattmann2023videoldm, blattmann2023stablevideodiffusionscaling} tasks. 
For face synthesis, a range of methods have emerged to tackle essential tasks, including 3D avatar creation~\cite{wang2022rodingenerativemodelsculpting, zhang2024rodinhd}, avatar reenactment~\cite{lan2023gaussian3diff,deng2024portrait4d, deng2024portrait4dv2} and texture map generation~\cite{Paraperas_2023_ICCV, galanakis2024fitdiffrobustmonocular3d}.
We adopt the concatenation strategy outlined in~\cite{Paraperas_2023_ICCV, galanakis2024fitdiffrobustmonocular3d} to generate novel viewpoint images and their corresponding shape normals simultaneously.
Based on Stable Diffusion~\cite{Rombach_2022_CVPR}, Arc2Face~\cite{deng2019arcface} and InstantID~\cite{wang2024instantid} generate facial images given an input subject. Notably, Arc2Face exhibits strong generalization abilities in facial image generation, leveraging an up-sampled subset of WebFace42M~\cite{Zhu_2021_CVPR}. 
Also, it introduces a robust identity conditioning mechanism, which is integrated into our method.
\vspace{-0.5cm}
\paragraph{Novel View Image Synthesis}
DreamFusion~\cite{poole2022dreamfusion} introduces the use of Score Distillation Sampling (SDS), incorporating a pre-trained text-to-image diffusion model~\cite{Rombach_2022_CVPR} alongside a NeRF model, to synthesize 3D objects from text prompts. 
In this way, its authors proved that they can efficiently generate 3D objects despite the lack of large-scale datasets; by exploiting the generalization abilities of an image generation network.
Although subsequent studies focus on better distillation strategies~\cite{zhou2024headstudio, raj2023dreambooth3d, Magic123}, the aforementioned approaches are time-consuming and require complex balancing and additional constraints~\cite{gao2024cat3d}.
Closer to our work, ID-to-3D~\cite{NEURIPS2024_b0171a59} and Arc2Avatar~\cite{gerogiannis2025arc2avatar}  use score distillation to generate 3D facial geometry, lacking though photorealistic results.
Zero123~\cite{liu2023zero1to3} introduces a multi-view diffusion model conditioned on a reference image and a camera pose. 
Following it, methods such as One-2-3-45~\cite{liu2023one2345}, SyncDreamer~\cite{liu2023syncdreamer}, Consistent1-to-3~\cite{weng2023consistent123} and Cascade-Zero123~\cite{Cascadezero123} further focus on multi-view consistency by introducing priors during the denoising process.
Studies such as Zero123++~\cite{shi2023zero123plus}, Era3D~\cite{li2024era3d} and Morphable Diffusion~\cite{chen2024morphable} generate fixed viewpoints given the input image, without being able to handle arbitrary views.
In Cat3D~\cite{gao2024cat3d}, a general multi-view diffusion model is introduced, using a powerful camera pose conditioning mechanism while using 3D attention layers.
Although it has been a robust method for novel view synthesis, it does not focus on facial novel view synthesis, 
lacks shape normal generation capabilities and is a closed source framework. 
DiffPortrait3D~\cite{Gu_2024_CVPR} showcases novel view capabilities, given an input facial image.
Unlike our method, which employs image-free camera conditioning, DiffPortrait3D employs an image-driven approach for the desired output viewpoints and can generate only near-front views.
More recent works~\cite{ taubner2024cap4dcreatinganimatable4d, prinzler2024joker} have explored diffusion-based frameworks for generating controllable and animatable avatars.
Concurrently to our approach, Pippo~\cite{Kant2025Pippo} and DiffPortrait360~\cite{gu2025diffportrait360consistentportraitdiffusion} demonstrated \allaround{} avatar generation, without, though, being able to sample novel identities. 
\vspace{-0.5cm}

\paragraph{Novel View Video Synthesis}
Recently, video generation models~\cite{girdhar2024emuvideofactorizingtexttovideo, blattmann2023videoldm, blattmann2023stablevideodiffusionscaling} have proven their ability to generate photorealistic models. 
Methods such as IM-3D~\cite{melaskyriazi2024im3diterativemultiviewdiffusion}, V3D~\cite{chen2024v3d} and SV3D~\cite{voleti2024sv3dnovelmultiviewsynthesis} integrate an off-the-shelf video diffusion model for generating novel viewpoints of a reference object.
Still, the video generation step makes these methodologies computationally demanding~\cite{gao2024cat3d}.
Additionally, they are often restricted by the requirement for specific camera trajectories, typically revolving around a central subject.
In contrast, our approach focuses on implementing multi-view diffusion networks that handle unordered camera poses.

\def \var {1}
\begin{figure*}[!ht]
\begin{center}
 \begin{subfigure}[b]{\var\textwidth}
      \includegraphics[width=\textwidth]{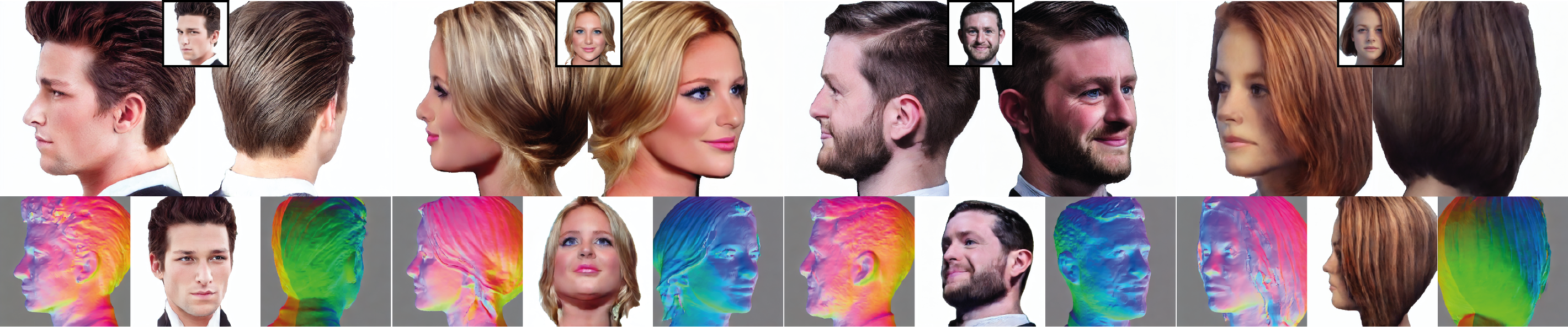}
 \end{subfigure}
 \caption{
    Samples generated by \ourname{} on ``in-the-wild'' images: Given only the input images (shown in a small box at the top center), our method produces high-fidelity images from novel angles, along with corresponding shape normals $\mathcal{N}$.}
\label{fig:results}
\end{center}
\vspace{-0.7cm}
\end{figure*}

\section{Method}
\ourname{} incorporates a multi-view diffusion model, trained on a 3D facial dataset, capable of generating a full head portrait of an input subject, given their identity embedding and a number of views (Fig.~\ref{fig:method}).
In the following sections, we first describe the proposed diffusion model, its training scheme, and the proposed sampling strategy given a single input view.

\subsection{Multi-View Diffusion Model}\label{sec:description}
\ourname{} employs a latent multi-view diffusion model alongside a powerful identity conditioning mechanism~\cite{paraperas2024arc2face}, enabling the generation of novel views of an input scene containing a person. 
Our approach leverages as conditioning inputs $M \in \{1,3\}$ pairs of sparse views of a subject, their respective shape normals and their associated camera poses.
Let $\mathbf{I}_i$ denote a picked conditioned view and $\bar{\mathbf{I}}_i$ a cropped and aligned version of $\mathbf{I}_i$. 
We utilize a pre-trained face recognition network (ArcFace~\cite{deng2019arcface}) $\phi$ to extract the identity embedding vector $\mathbf{w} = \phi(\bar{\mathbf{I}}_i) \in \mathbb{R}^{512}$ capable of incorporating crucial identity features.
Then, we inject the identity information into the diffusion model as proposed in Arc2Face~\cite{deng2019arcface}. 
In particular, a text prompt of ``\emph{a photo of $<$id$>$ person}, is fed to a CLIP text encoder~\cite{Radford2021LearningTV} followed by the tokenization step.
Followed by padding, the identity embedding vector $\mathbf{w}$ replaces the corresponding $<$id$>$ token  
resulting in $\mathbf{s} = \{e_1, e_2, e_3, \hat{\mathbf{w}}, e_5\}$, where $\hat{\mathbf{w}} \in \mathbb{R}^{768}$ is the padded version of $\mathbf{w}$.
The vector $\mathbf{s}$ is fed to the encoder $\mathcal{C}$, and finally we retrieve the corresponding conditioning vector $\mathbf{c} = \mathcal{C}(\mathbf{s}) \in \mathbb{R}^{N\times 768}$, where $N$ is the maximum sequence length.

The proposed latent multi-view UNet~\cite{DBLP:journals/corr/RonnebergerFB15} is similar to the one introduced in Cat3D~\cite{gao2024cat3d}.
Our diffusion model is designed to concurrently generate $P = (M + K) = 8$ pairs of facial images and their corresponding normals $\mathcal{N}$, given an input conditioning vector $\mathbf{w}$, a number of $M$ views and the $P$ camera poses.
Let $I_i^{cond}$ represent the input human face images, $I_j^{tgt}$ the target face images from novel viewpoints, and $\mathcal{N}_i^{cond}$, $\mathcal{N}_j^{tgt}$ their respective shape normals, where
$i\in \{ 1,2, ..., M \}$, $j \in \{ 1,2, ..., N\} $ with $M$ being the maximum number of the input conditional views and $N$ the number of target viewpoints.
Following~\cite{Paraperas_2023_ICCV, gao2024cat3d, galanakis2024fitdiffrobustmonocular3d}, we employ a pre-trained AutoEncoder of Stable Diffusion 1.5 (SD 1.5)~\cite{Rombach_2022_CVPR}, consisting of an encoder $\mathcal{E}$ and a decoder $\mathcal{D}$.
For each input conditioning image, we first obtain the corresponding latent feature maps: $\mathbf{z}_{I_i^{cond}}= \mathcal{E}(I_i^{cond}) \in \mathbb{R}^{4\times 64 \times 64}$ and $\mathbf{z}_{\mathcal{N}_i^{cond}} = \mathcal{E}(\mathcal{N}_i^{cond}) \in \mathbb{R}^{4\times 64 \times 64}$. These are then concatenated channel-wise. 
Similarly, we apply the same procedure to extract the latent vectors for the target viewpoints.
The corresponding camera pose information is incorporated for both conditioning and target viewpoints using the mechanism proposed in Cat3D~\cite{gao2024cat3d}.
For each of the $P$ views, the latent feature maps $\mathbf{z}_i$ are concatenated channel-wise with the respective ray representation maps~\cite{gao2024cat3d,srt22} $\mathbf{r}_i^{cond}, \mathbf{r}_j^{tgt} \in \mathbb{R}^{149\times 64 \times 64}$, which encode the ray origin and direction.
A binary mask $\mathbf{m} \in \{ \mathbf{0}, \mathbf{1} \}^{1 \times 64 \times 64}$ is then appended to differentiate between conditional and target latent vectors. 
Finally, we retrieve the conditioning and target latent feature maps $\bar{\mathbf{z}}_i^{cond} = \{ \mathbf{z}_{I_i^{cond}}, \mathbf{z}_{\mathcal{N}_i^{cond}} \mathbf{r}_j^{cond} \mathbf{m}_j^{cond}\}$ 
and $\bar{\mathbf{z}}_i^{tgt} = \{ \mathbf{z}_{I_i^{tgt}}, \mathbf{z}_{\mathcal{N}_i^{tgt}} \mathbf{r}_j^{tgt}, \mathbf{m}_j^{tgt} \}$.

As proposed in video diffusion models~\cite{blattmann2023stablevideodiffusionscaling}, we initialize our method using a pre-trained LDM model (Arc2Face~\cite{paraperas2024arc2face}) trained on large-scale datasets while adding additional attention layers, connecting the multiple latent feature maps. 
As in ~\cite{gao2024cat3d}, we integrate 3D attention layers~\cite{shi2023MVDream} by adding them between the original 2D self-attention layers of the LDM, and we fine-tune all layers of the multi-view UNet for improved multi-view consistency.

\subsection{Training Details}\label{sec:training}
We begin from the publicly available \emph{Arc2Face}~\cite{paraperas2024arc2face}, and we adhere to the training schemes introduced in~\cite{blattmann2023stablevideodiffusionscaling, gao2024cat3d}.
Specifically, we adapt the EDM framework~\cite{karras2022edm} to Arc2Face, training it for 31,000 iterations on its dedicated dataset.
To accommodate the input latent feature maps, we expand the input and output convolutional layers channels, initializing these by copying the existing weights to the shape-normal channels while randomly initializing the camera-pose dimensions. 
As in ~\cite{gao2024cat3d}, we shift the log-to-signal ratio by log(N), where N is the number of the target images (N=7). 
We randomly select a conditioning view that includes part of the frontal face during each training iteration, as required for the identity embedding extraction. 
We then randomly pick the N target images and calculate the relative camera angles.
To enhance the dataset, we replace the white background with a random color, in 50$\%$ of the samples.
All the training viewpoints are encoded using the encoder $\mathcal{E}$, with noise added only to the target latent vectors, while conditioning vectors remain unchanged.
Following the classifier-free guidance (CFG) training scheme~\cite{ho2022classifier}, with a probability of $\mathcal{P}_{uncond}=0.15$, we randomly replace the identity vector with the empty string, and the conditioning images with zero-ed ones.
We first train the model conditioned on a single view for 600k iterations.
Then, for the additional 1M iterations,  we vary the conditioning views by randomly choosing 0, 1 or 3 conditioning views, corresponding to 8, 7 and 5 target views, each with a probability of $\mathcal{P}_=1/3$.

\subsection{Novel view sampling}\label{sec:sampling}

\ourname{} synthesizes novel views for an input subject $\mathbf{I}$, given a number of views.
In this section, we introduce a robust sampling strategy designed to produce a large number of consistent views that comprehensively cover a full head, given only a single input image.
Achieving consistency across viewpoints requires a carefully structured camera pose selection order. 
Therefore, we employ a three-step sampling process:
a) aligning input views and extracting the corresponding shape normals,
b) generating anchor images that provide complete coverage of the \allaround{} human head and
c) synthesizing intermediate views by leveraging both the input views and the closest anchor images.

\paragraph{Alignment and Shape Normals generation}
Given the input image $\mathbf{I}$, we extract its identity embedding $\mathbf{w}$ as described in Sec.~\ref{sec:description}. 
Then, we obtain $\hat{\mathbf{I}}$, a cropped and aligned version of $\mathbf{I}$, using the alignment procedure presented in Panohead~\cite{an2023panohead}.
To generate the shape normals $\mathcal{N}$, we treat this as an in-painting task, thus retrieving the normals through the conditional guidance sampling approach described in Relightify~\cite{Paraperas_2023_ICCV}. 
Specifically, using the aligned image $\hat{\mathbf{I}}$ and its identity embedding vector $\mathbf{w}$, we employ a binary visibility mask $m$, marking only the image channels as visible and setting the shape normal channels as non-visible.
By applying the ``channel-wise in-painting'' algorithm, the corresponding shape normals are retrieved.
This process uses the EDM sampler presented in ~\cite{karras2022edm} alongside the DDPM~\cite{ho2020denoising} discretization steps and runs for 50 steps.
A more detailed presentation of this approach is presented in the supplemental material.

\paragraph{Generating anchor and intermediate views}
\ourname{} can generate any arbitrary viewpoint, given the input aligned facial image $\hat{\mathbf{I}}$ and the corresponding shape normals $\mathcal{N}$.
However, since it was trained to generate only a limited number of views per sampling process, a carefully designed sampling strategy is essential to produce a wide range of output views.
Since the target subject is centered in the scene, we first generate $M = 7$ anchor images $\mathbf{A}_i$ and corresponding anchor shape normals $\mathbf{A}_{\mathcal{N}_i},~i\in \{ 1, ..., M \}$, covering a \allaround{} angle range of the subject ($\pm 45^{\circ}, \pm 90^{\circ}, \pm 135^{\circ}, 180^{\circ}$), as proposed in ~\cite{gao2024cat3d}. 
Using these anchors, any number of intermediate views can then be synthesized by conditioning on image triplets $\{ (\mathbf{I},  \mathcal{N}), (\mathbf{A}_k, \mathbf{A}_{\mathcal{N}_k}), (\mathbf{A}_l, \mathbf{A}_{\mathcal{N}_l}) \}$, where $k,l$ where represent the closest anchor images. 
 This approach ensures that each intermediate view remains consistent with both the input aligned image $\hat{\mathbf{I}}$, and the closest already generated views.
For both sampling processes, we use the EDM sampler facilitated by the EDM discretization steps~\cite{karras2022edm}, while it runs for 50 steps with a guidance scale set to 3.
This method enables the generation of 48, 88, or more novel views for a single input subject $\mathbf{I}$, depending on the chosen angle step, thus covering the entire scene.

\begin{figure*}[!ht]
\begin{center}
 \begin{subfigure}[b]{\var\textwidth}
      \includegraphics[width=\textwidth]{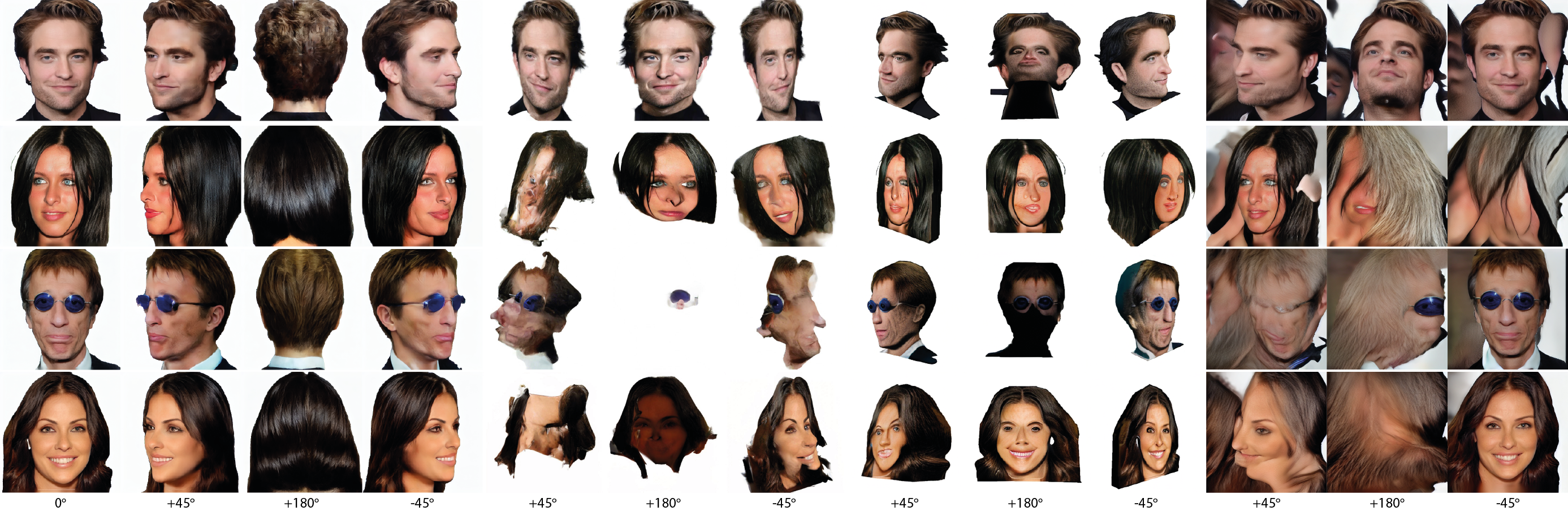}
 \end{subfigure}
  \begin{scriptsize}
             \makebox[0.10\linewidth][c]{\textbf{(a)} Input}\hfill
             \makebox[0.2\linewidth][c]{\textbf{(b)} \ourname{} (Ours)}\hfill
             \makebox[0.2\linewidth][c]{\textbf{(c)} SV3D~\cite{voleti2024sv3dnovelmultiviewsynthesis}}\hfill
             \makebox[0.2\linewidth][c]{\textbf{(d)} Zero123-XL~\cite{objaverseXL}}\hfill
             \makebox[0.2\linewidth][c]{\textbf{(d)} DiffPortrait3d~\cite{Gu_2024_CVPR}}\hfill
         \end{scriptsize}
 \caption{Qualitative comparison between \ourname{}, Zero123-XL~\cite{objaverseXL}, SV3D~\cite{voleti2024sv3dnovelmultiviewsynthesis} and DiffPortrait3D~\cite{Gu_2024_CVPR} at angles \{$\pm 45^{\circ}, +180^{\circ}$\}.
 It is clear that \ourname{} effectively generates high-quality novel views from ``in-the-wild'' input images, whilst SV3D produces distorted outputs, Zero123-XL generates unnaturally squared avatars and DiffPortrait3D cannot handle such angles.
 }
 \vspace{-0.7cm}
\label{fig:qual_comparison}
\end{center}
\end{figure*}

\section{Experiments} \label{sec_experiments}

\subsection{Training Dataset} \label{sec:training_datase}
Training such a model requires a large multi-view dataset containing a great number of subjects $\mathcal{S}$.
For each person $\mathcal{S}_i$, it is necessary to acquire a set of images $\mathbf{I}_k^i$, their corresponding shape normal maps $\mathcal{N}_k^i$, camera poses $\mathcal{C}_k^i$ alongside with their corresponding identity embedding vector $\mathbf{w}^i$, where $k \in \{1, 2, ..., N_i \}$ and $N_i$ is the number of the available views for the $i$-th scene.
Due to the lack of such a dataset, we create a large-scale synthetic dataset using the publicly available Panohead~\cite{an2023panohead}.
We first sample $\sim$10k subjects and manually remove instances with artifacts in the back of the head, resulting in $\sim$7k distinct identities. 
We render images from 125 different viewpoints for each person to cover the entire head.
Simultaneously, we obtain each subject's facial shape by extracting their opacity values from the triplane feature maps and then applying the marching cubes~\cite{conf/siggraph/LorensenC87} algorithm.
We render their respective normal maps using the acquired facial shape using Pytorch3d~\cite{ravi2020pytorch3d}.
All the images depicting a frontal head are fed to the Arc-Face~\cite{deng2019arcface} network to extract their corresponding identity vectors $\mathbf{w}$.
All in all, we end up with a synthetic dataset containing about 7k individuals, rendered from $N=125$ different angles, the respective shape normals, camera poses, and their identity embedding vector $\mathbf{w}$.

\subsection{Qualitative Comparisons}
\subsubsection{Novel View Synthesis comparisons}
In this section, we present a qualitative comparison between  our model and other state-of-the-art multi-view diffusion models, SV3D~\cite{voleti2024sv3dnovelmultiviewsynthesis} 
Zero123-XL~\cite{objaverseXL} and DiffPortrait3D~\cite{Gu_2024_CVPR}, focusing on their output under $\pm 45^{\circ}, +180^{\circ}$ angles.
SV3D, a latent video diffusion model, creates \allaround{} videos from a single input image. 
Zero123-XL, a multi-view diffusion model, synthesizes novel perspectives based on the input image and specified camera poses.
DiffPortrait3D~\cite{Gu_2024_CVPR} generates novel views by conditioning the desired camera on an input image and refining the initial noise through a fine-tuning step to achieve optimal results.
Figure \ref{fig:qual_comparison} demonstrates scenarios where the input ``in-the-wild'' images (Fig.~\ref{fig:qual_comparison}a ) are fed to these 4 different networks and their respective generations under $\pm 45^{\circ}, +180^{\circ}$ viewpoints.
As illustrated, \ourname{} efficiently generates accurate novel viewpoints of the input subject, while SV3D distorts the input image in its generated views, Zero123-XL  produces unnatural, square-like facial avatars and DiffPortrait3D cannot handle those angles at all.

\begin{figure}[!h]
    \centering
    \includegraphics[width=\linewidth]{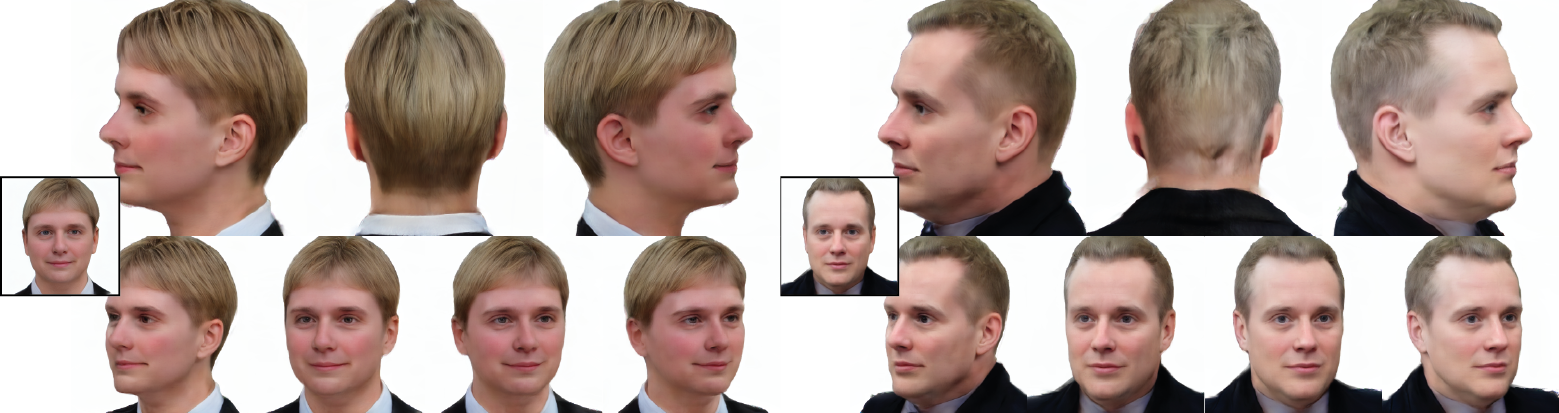}
    \caption{Given the input identities (shown in the small boxes), we present the results of novel view synthesis after applying 3D Gaussian splatting~\cite{kerbl3Dgaussians} to the views generated by our model.}
    \label{fig:gs}
\end{figure}

\subsection{3D Reconstruction}
To evaluate our model’s consistency in generating novel views from an input image, we assess its ability to reconstruct an input identity through Gaussian Splatting (3DGS)~\cite{kerbl3Dgaussians}.
More specifically, given an input identity, we apply the sampling strategy outlined in Section~\ref{sec:sampling} to generate 48 novel views.
These views are then used to reconstruct the identity through Gaussian Splatting.
As in ~\cite{gao2024cat3d}, we modify the provided 3DGS code to incorporate the  LPIPS~\cite{zhang2018perceptual} loss between the ground-truth and the generated viewpoints.
In this way, we can deal with small inconsistencies between nearby viewpoints. 
We present the novel view synthesis of two input identities (small squares) in Figure~\ref{fig:gs}, which clearly illustrates that our proposed methodology can generate consistent subjects.

\begin{figure}[h]
\centering
\includegraphics[width=\linewidth]{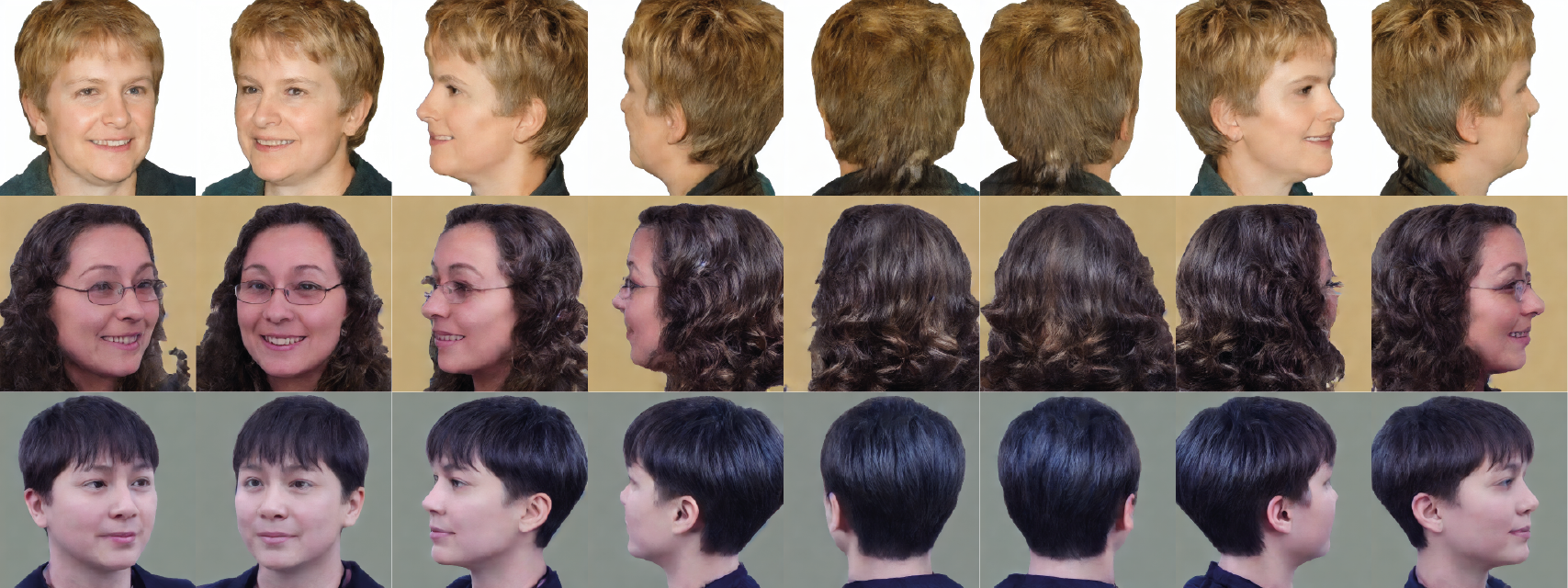}
    \caption{Samples generated using unconditional sampling. \ourname{} can generate novel identities without any prior input.}
    \label{fig:unconditional}
\end{figure}

\subsection{Unconditional Sampling}
\ourname{} is trained following the Classifier-Free Guidance training scheme. Thus, our architecture can generate novel multi-view identities without any prior conditioning. 
By setting the input identity embedding equal to the empty string, our model can generate a novel identity along with different viewpoints of that identity.
Figure~\ref{fig:unconditional} presents some examples of those identities.

\subsection{Quantitative Comparisons}

We compare our method's ability to generate novel views with the current state-of-the-art models: a) Eg3D~\cite{Chan_2022_CVPR} and 
Panohead~\cite{an2023panohead} which are NeRF-based approaches generating frontal and full-head portraits respectively, b) Zero123~\cite{Cascadezero123}, Zero123-XL~\cite{objaverseXL} and DiffPortrai3D~\cite{Gu_2024_CVPR}, which are multi-view diffusion models and c) SV3D~\cite{voleti2024sv3dnovelmultiviewsynthesis}, a multi-view video diffusion model.
For SV3D, we selected the SV3D-p variant, as it supports flexible viewing angles, whereas SV3D-u operates only with fixed viewpoints.
We validate our approach using the NeRSemble dataset~\cite{kirschstein2023nersemble}, which includes 222 unique identities recorded from 16 angles. 
We randomly select a timestamp from one of their video sequences for each individual and extract a random subset of views. 
All selected frames are then centered, and one of these views is used as input for the tested methods.
We evaluate their reconstruction performance using L2 distance, LPIPS~\cite{zhang2018perceptual}, SSIM, and an Identity Similarity score. 
To calculate the Identity Similarity score, we feed both the ground-truth and the reconstructed images into the ArcFace~\cite{deng2019arcface} and VGGFace~\cite{vggface} networks  and measure the cosine similarity between their final feature vectors.

Table~\ref{tab:nersemble} summarizes the reconstruction performance of each network.
\ourname{}, achieves state-of-the-art results across all cases when evaluated on the LPIPS, SSIM, and Identity Similarity metrics. 
Additionally, it outperforms all diffusion-based approaches in terms of L2 distance. Although our method trails slightly behind NeRF-based approaches, this is expected given the inherent advantages of NeRF-based models in novel view synthesis tasks.
Notably, methods like Eg3D and Panohead require a time-intensive fitting process, which can occasionally fail. In contrast, our proposed approach relies solely on an efficient sampling process, eliminating the need for fitting.

\begin{table}[!t]
\begin{tiny}
\setlength{\tabcolsep}{5pt}
\begin{center}
\begin{tabular}{llccccc}
\toprule
    \multicolumn{2}{c}{\textbf{Method}} & $\textbf{L2}\downarrow$  & \textbf{LPIPS}$\downarrow$ & \textbf{SSIM}$\uparrow$ & \textbf{ID Sim~\cite{deng2019arcface}} $\uparrow$  & \textbf{ID Sim~\cite{vggface}} $\uparrow$\\
\midrule
\multirowcell{2}{NeRF\\based} & Eg3d~\cite{Chan_2022_CVPR} &  0.025 & 0.4 & 0.55 & 0.31 & 0.89 \\
   & Panohead~\cite{an2023panohead} & \textbf{0.012} & 0.32 & 0.65 & 0.27 & 0.88\\
\hline
\multirowcell{4}{Diffusion\\based}   & Zero123~\cite{liu2023zero1to3} & 0.195  & 0.515   & 0.55 & 0.169 & 0.44  \\
   & Zero123-XL~\cite{objaverseXL} & 0.198  &  0.51   & 0.563  & 0.118 & 0.442\\
   & SV3D~\cite{voleti2024sv3dnovelmultiviewsynthesis} & 0.087 & 0.41 &  0.660  &  0.36 & 0.881\\
   & DiffPortrait3d~\cite{Gu_2024_CVPR} & 0.1 & 0.5 &  0.35  &  0.55 & 0.887\\
\hline
   & \ourname{}~(Ours) & 0.033 & \textbf{0.3} & \textbf{0.73} & \textbf{0.61} & \textbf{0.911} \\
\bottomrule    
\end{tabular}
\captionof{table}{Reconstruction performance on the NeRSemble~\cite{kirschstein2023nersemble} dataset shows that \ourname{} achieves state-of-the-art results in LPIPS, SSIM and ID Sim metrics while performing on par with the leading models in terms of L2 distance.}
\label{tab:nersemble}
\vspace{-0.65cm}
\end{center}
\end{tiny}
\end{table}

\section{Ablation Studies}\label{sec:ablation}

\subsection{Component analysis}
In this section, we conduct ablation studies for the importance of the identity mechanism, the shape normals and the input ground truth image.
For this reason, we trained the following three models while using the same training data:
a) our proposed architecture, without integrating the input conditioning view,
b) a Stable Diffusion-based model without integrating the robust identity mechanism, and
c) our model without generating any shape normals.

\paragraph{Use of input image:}
Starting from a pre-trained Arc2Face model using the EDM framework, 
we trained a multi-view model to generate novel viewpoints given only an input identity embedding vector. 
Notably, \ourname{} has also been trained to be able to generate input identities without providing the conditioning view. 
In this case, it generates subjects close to the input face, as illustrated in Figure~\ref{fig:ablation-sampling}.

\begin{table}[!t]
\centering
\setlength{\tabcolsep}{1.5pt}
\scriptsize
\begin{center}
\begin{tabular}{lcccccc}
\toprule
 & \textbf{L2} $\downarrow$ & \textbf{LPIPS} $\downarrow$ & \textbf{SSIM} $\uparrow$ \\
 \midrule
 \ourname{} (w/o Input Image) & 0.1246 & 0.4299 & 0.568 \\
 \ourname{} (w/o identity embedding) & 0.028 & 0.26 & 0.70    \\
 \ourname{} (w/o Normals)  & 0.056 & 0.32 & 0.65 \\
 \ourname{} & \textbf{0.018 }& \textbf{0.22} & \textbf{0.75}   \\
 
\bottomrule
\end{tabular}
\captionof{table}{Ablation study: We evaluate the performance of \ourname{} alongside three variations, demonstrating that the proposed architecture achieves the highest performance across all reconstruction metrics.
}
\label{tab:ablation}
\end{center}
\end{table}

\def \var {1}
\begin{figure}[!t]
    \centering
    \begin{subfigure}[t]{\var\linewidth}
        \includegraphics[width=\linewidth]{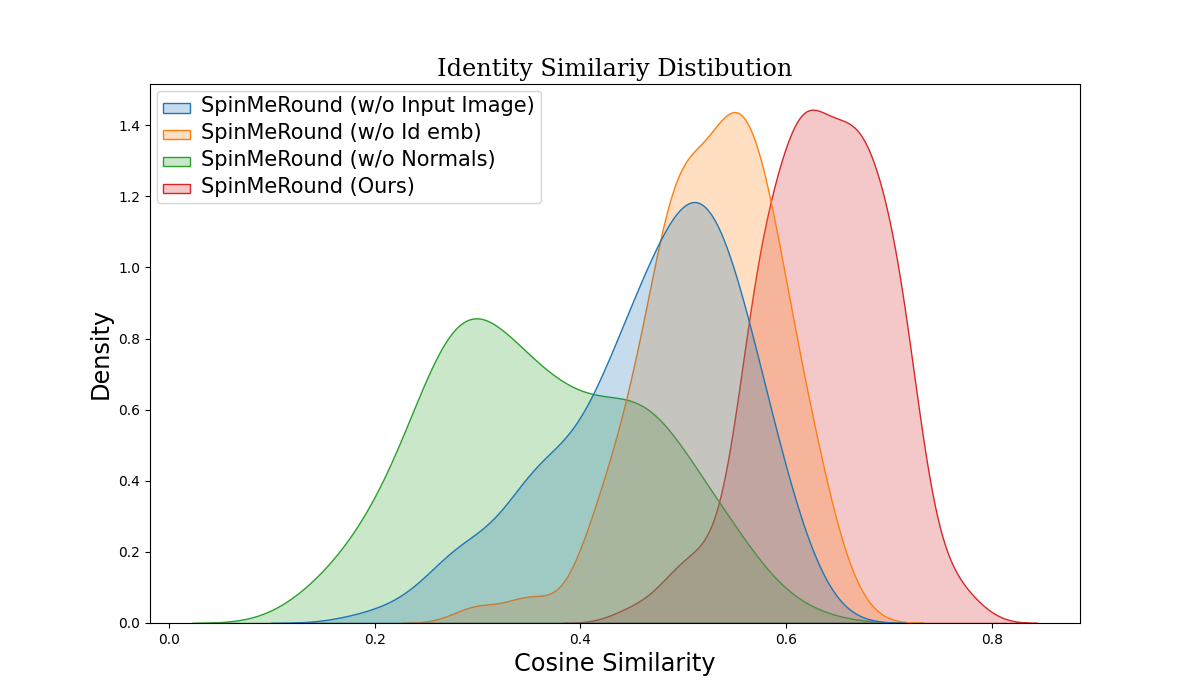}
    \end{subfigure}
    \caption{
    Ablation Study: We compare the identity similarity between four models: \ourname{} (Ours), a similar model without the input image, ones without the identity embedding mechanism, and one without generating shape normals $\mathcal{N}$.
Results show that our proposed architecture achieves the highest identity similarity.}
    \label{fig:ablation}
    \vspace{-0.5cm}
\end{figure}

\begin{figure*}[t!]
    \centering
    \begin{subfigure}[b]{\linewidth}
        \includegraphics[width=\linewidth]{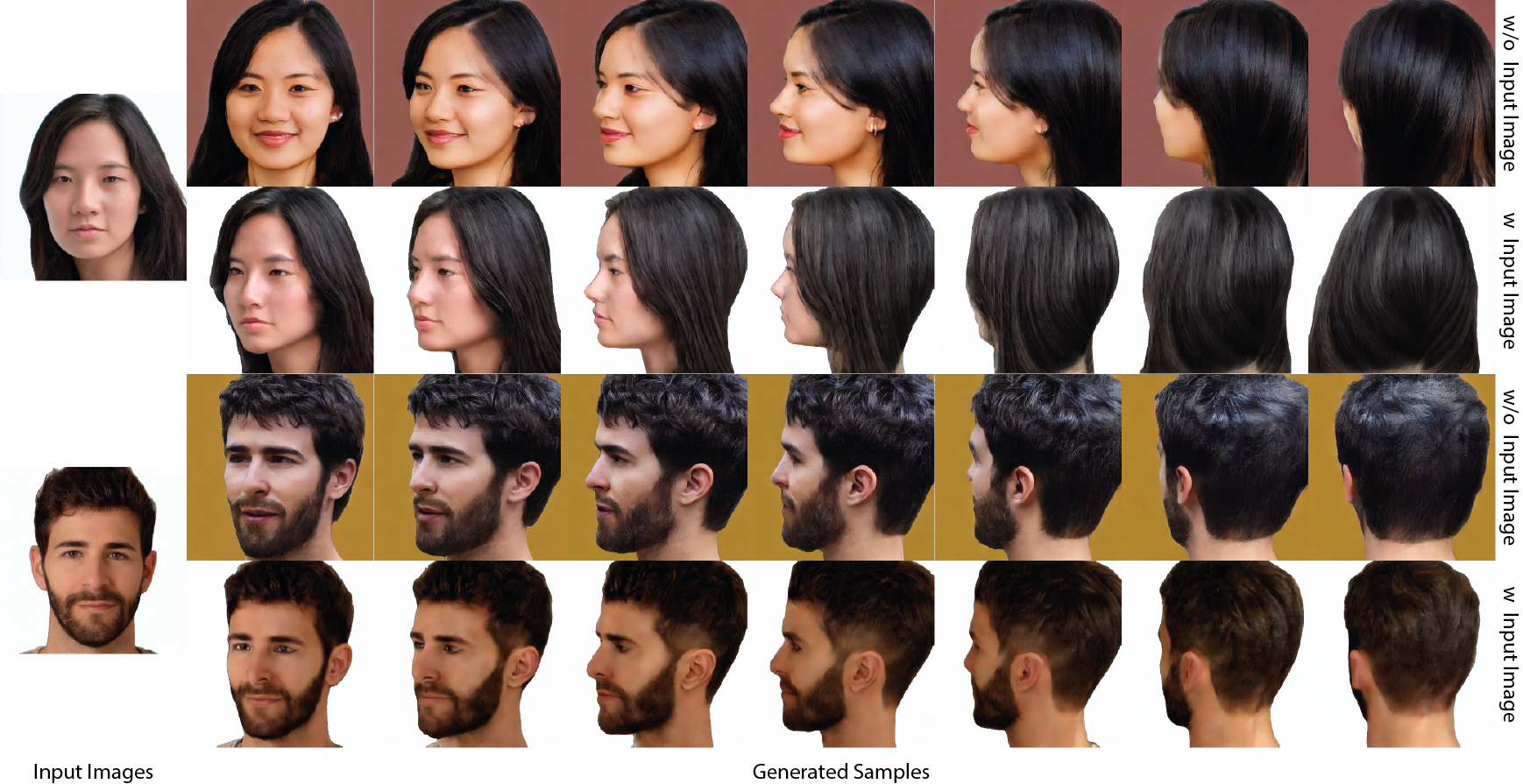}
    \end{subfigure}
    \caption{From the input images on the left, we present samples generated by our method using only the corresponding identity embedding vectors (top row) and using also the input image (bottom row). The resulting subjects closely resemble the input identities.}
    \label{fig:ablation-sampling}
    \vspace{-0.3cm}
\end{figure*}

\begin{figure}[t!]
    \centering
    \begin{subfigure}[b]{\linewidth}
        \includegraphics[width=\linewidth]{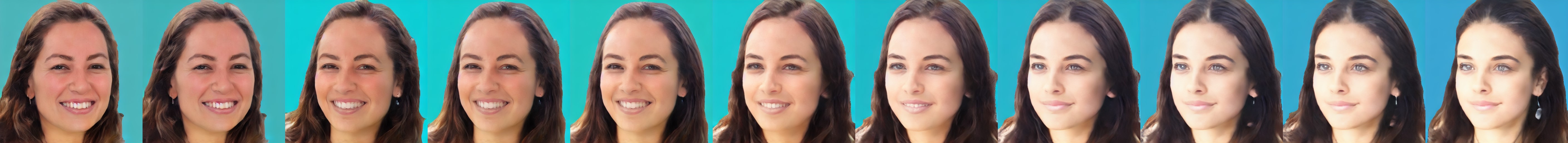}
    \end{subfigure}
    \begin{subfigure}[b]{\linewidth}
        \includegraphics[width=\linewidth]{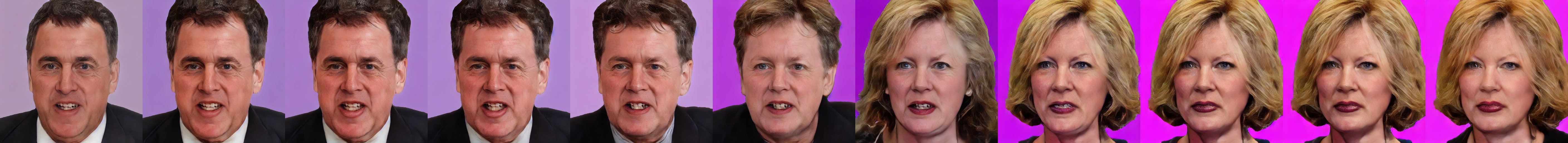}
    \end{subfigure}
    \begin{subfigure}[b]{\linewidth}
        \includegraphics[width=\linewidth]{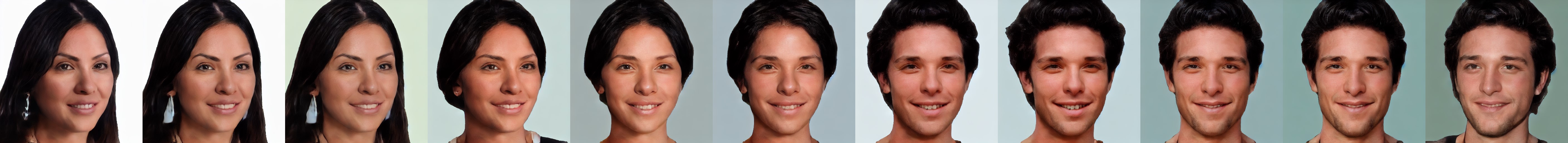}
    \end{subfigure}
    \begin{subfigure}[b]{\linewidth}
        \includegraphics[width=\linewidth]{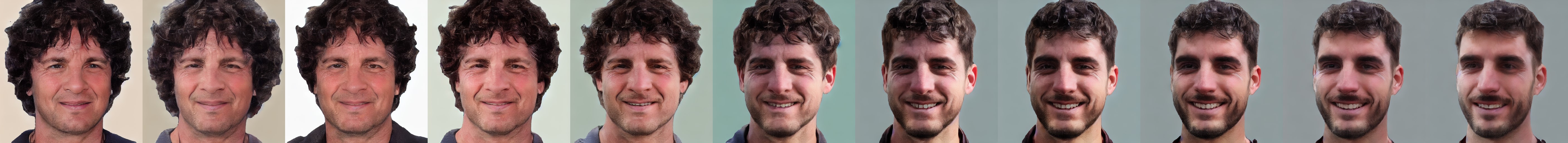}
    \end{subfigure}
    \caption{Identity interpolation between subject pairs, showcasing our method generates smooth transitions between identities. }
    \label{fig:interpolation}
    \vspace{-0.5cm}
\end{figure}

\paragraph{Identity embedding mechanism:}
In this case, we start from a pre-trained Stable diffusion 1.5 architecture trained using the EDM framework.
This model gets trained following the same training parameters, as presented in Sec~\ref{sec:training}.
Instead of using the proposed identity conditioning mechanism, we feed an empty string in the conditioning attention layers of the denoising UNet.

\paragraph{Use of Normals:}
We also train a variant of the network similar to \ourname{}, but without generating shape normals. 
This version can only generate novel views without having any additional information about the facial shape.

Aiming to showcase the importance of each component, we conducted a multi-view reconstruction experiment. 
More specifically, we sample 100 distinct identities using Panohead~\cite{an2023panohead}.
For each subject, we sample 10 different views along with an input frontal view.
Given the frontal view as input to each separate model, we reconstruct the remaining views.
We measure the performance of each model by calculating their discrepancy using L2 distance, LPIPS~\cite{zhang2018perceptual}, SSIM and ID Similarity.
As presented in Table~\ref{tab:ablation} and Figure~\ref{fig:ablation}, our proposed methodology performs better regarding all reconstruction metrics while achieving the highest identity similarity score.

\subsection{Importance of ID features}
Choosing to condition the stable diffusion model in identity embeddings is an important design choice for our network.
Those forms of representation contain compact information extracted from a FR model (ArcFace~\cite{deng2019arcface}), trained in millions of different images and subjects.
Those features allow linear interpolation between the facial characteristics of different subjects.
Hence, while using only the identity embedding layer, we interpolate between 2 distinct identities and present their linear interpolation results in Figure~\ref{fig:interpolation}. It is clearly shown that our model generates smooth transitions between the generated identities.

\section{Conclusion}

In this paper, we presented \ourname{}, a multi-view latent diffusion model, which generates all-around head portraits of an input subject given a number of input views alongside an input identity embedding. 
Additionally, we introduced a sampling strategy for generating consistent intermediate views, given an unconstrained input ``in-the-wild'' facial image.
Being trained sorely on multi-view synthetic data, we showcase our method's abilities by beating the current state-of-the-art multi-view diffusion models in novel view synthesis experiments.

\noindent\textbf{Acknowledgments:}
S. Zafeiriou and part of the research was funded by the EPSRC Fellowship DEFORM (EP/S010203/1) and EPSRC Project GNOMON (EP/X011364/1).
The authors gratefully acknowledge the scientific support and HPC resources provided by the Erlangen National High Performance Computing Center (NHR@FAU) of FAU under the NHR projects b143dc and b180dc. NHR funding is provided by federal and Bavarian state authorities. NHR@FAU hardware is partially funded by the German Research Foundation (DFG) – 440719683. Additional support was also received by the ERC - project MIA-NORMAL 101083647, DFG 513220538, 512819079, and by the state of Bavaria (HTA).

{
    \small
    \bibliographystyle{ieeenat_fullname}
    \bibliography{main}
}

\setcounter{section}{0}

\section{Limitations and Future Work}
Although \ourname{} showcases high-fidelity results, it has some limitations. 
More specifically, our model takes over structural limitations presented in Panohead~\cite{an2023panohead} due to the synthetic dataset used. 
This means that there are some inconsistencies in the generated eyes, hair and noses.
Moreover, using the alignment pipeline sometimes results in failure cases, due to misalignment.

All in all, the fact that we do not use any captured data limits our model's capabilities.
Hence, this is a direction that we plan to explore in future work. 
Captured datasets such as FaceScape~\cite{zhu2023facescape}, Renderme-360~\cite{2023renderme360} and NeRSemble~\cite{kirschstein2023nersemble} can be used to further improve our results.
Finally, integrating video diffusion models~\cite{blattmann2023stablevideodiffusionscaling} can be another direction for our future work, to improve the consistency of the generated viewpoints.

\section{Training Details}
\ourname{} begins training using the publicly available Arc2Face model ~\cite{paraperas2024arc2face}.
The Arc2Face model is built upon \emph{Stable Diffusion 1.5}~\cite{Rombach_2022_CVPR}, meaning that it incorporates the following preconditioning functions, according to the EDM framework~\cite{karras2022edm}:
\begin{equation*}
\begin{aligned}
    &c_{skip}^{SD1.5}(\sigma )= 1, &c_{out}^{SD1.5}(\sigma )= -\sigma,\\
    &c_{in}^{SD1.5} =\frac{1}{\sqrt{\sigma^2+1}}, 
    &c_{noise}^{SD1.5}(\sigma ) = \argmax_{j \in [1000]} (\sigma - \sigma_j)
    \end{aligned}
\end{equation*}
As proposed in \cite{karras2022edm}, we modify the aforementioned pre-conditioning by:
\begin{equation*}
\begin{aligned}
    &c_{skip}(\sigma )= (\sigma^2 + 1), 
    &c_{out}(\sigma )= \frac{-\sigma}{\sqrt{\sigma^2 + 1}},\\
    &c_{in} =\frac{1}{\sqrt{\sigma^2+1}}, 
    &c_{noise}(\sigma ) = 0.25 \log \sigma,
    \end{aligned}
\end{equation*}
Furthermore, we use the proposed noise distribution and weighting functions $log\sigma \sim \mathcal{N}(P_{mean}, P^2_{std})$ and $\lambda(\sigma ) = ( 1 + \sigma^2)\sigma^{-2}$, with $P_{mean} = 0.7$ and $P_{std} = 1.6$.
We finetune the pre-trained Arc2Face model for 31k iterations, using the training dataset provided by the Arc2Face authors.

\subsection{Shape Normals Retrieving}

\begin{algorithm}[!h]
\caption{Shape Normals sampling using Guidance}\label{alg:shapenormals}
\begin{algorithmic}[1]
\Require The aligned facial ``in-the-wild'' image $\bar{\mathbf{I}}$, the gradient scale $\alpha$, 
the binary visibility mask $m$, the conditioning mechanism $\mathcal{C}$, and encoder $\mathcal{E}$.

\State $\mathbf{c} \leftarrow \mathcal{C}(\bar{\mathbf{I}})$, $\mathbf{z}_{gt} \leftarrow \{ \mathcal{E}(\bar{\mathbf{I}}) | \mathbf{0} \}$
\State $\mathbf{z}_0 \sim \mathcal{N}(\mathbf{0}, t_0^2\mathbf{I})$

\ForEach{ i \textbf{from} 0 \textbf{to} N-1} 
\State $\mathbf{\epsilon}_i \sim \mathcal{N}(\mathbf{0}, S_{noise}^2 \mathbf{I})$
\State $\gamma_i = \left\{
\begin{array}{cc}
 \min ( \frac{S_{churn}}{N}, \sqrt{2} - 1 ) & \text{if }t_i\in[S_{tmin},S_{tmax} ] \\
 0  & \text{otherwise} \\
\end{array} \right\} $
\State $\hat{t}_i \leftarrow t_i + \gamma_i t_i$
\State $\hat{\mathbf{x}}_i \leftarrow \mathbf{x}_i + \sqrt{\hat{t}_i^2 - t_i^2} \epsilon$
\State $\mathcal{L} \leftarrow  || (\mathbf{z}_{gt} - D_{\theta} (\hat{\mathbf{x}}_i ; \hat{t}_i, \mathbf{c} ) ) \odot m ||_2^2$
\State $\mathbf{d}_i \leftarrow (\hat{\mathbf{x}}_i - D_{\theta} (\hat{\mathbf{x}}_i ; \hat{t}_i, \mathbf{c}) - \alpha \frac{\partial \mathcal{L}}{\partial \hat{\mathbf{x}}_i} ) / \hat{t}_i$
\State $\mathbf{x}_{i+1} \leftarrow \hat{\mathbf{x}}_i + ( t_{i+1} - \hat{t}_i) \mathbf{d}_i$
\EndFor \\
\Return $\mathbf{z}_N$
\end{algorithmic}
\end{algorithm}

As mentioned in Section 3.3, given an input ``in-the-wild'' facial image, we first extract the respective shape normals $\mathcal{N}$.
Our proposed sampling methodology is presented in Algorithm ~\ref{alg:shapenormals} and is inspired from Relightify~\cite{Paraperas_2023_ICCV}. 
Given an aligned ``in-the-wild'' image, we follow the sampling algorithm presented in Algorithm ~\ref{alg:shapenormals}, where
$\odot$ denotes the Hadamard product $\bar{\mathbf{I}}$.
We guide the sampling process to generate the respective shape normals, based on the distribution of the training data.
In detail, we firstly extract the conditioning label, as described in Section 3.1, and the latent feature maps of the image $\bar{\mathbf{I}}$, which gets padded. 
After that, we sample the input gaussian noise.
For each sampling step, we estimate $\hat{x}_i$ as presented in steps 4, 5, 6 and 7.
Then, we compute the guidance loss by calculating the masked \emph{L2}-distance between the ground-truth latent vector $\mathbf{z}_{gt}$ and the estimated $D_{\theta} (\hat{\mathbf{x}}_i ; \hat{t}_i, \mathbf{c} )$.
We calculate the Euler step from $\hat{t}_i$ to $t_{i+1}$ by applying the formula in line 9. 
During sampling we set the guidance scale equal with $10^5$ and we run for $t=50$ sampling steps.
The sampling process takes about 2.4 seconds while it runs on an NVIDIA A100-PCIE.

\begin{figure}[t]
  \centering
  \includegraphics[width=\linewidth]{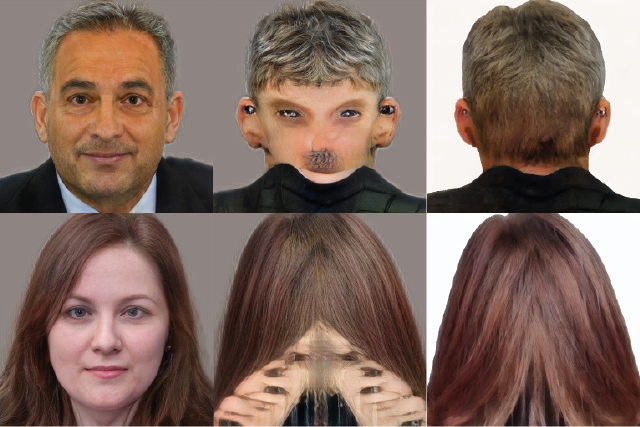}
     \makebox[0.3\linewidth][c]{Input}\hfill
     \makebox[0.3\linewidth][c]{Panohead~\cite{an2023panohead}}\hfill          
     \makebox[0.3\linewidth][c]{Ours}\hfill
  \caption{We compare the generated backhead between \ourname{}(Ours) and Panohead~\cite{an2023panohead}.}  \label{fig:panohead}    
\end{figure}

\begin{figure}[t]
  \centering
  \includegraphics[width=\linewidth]{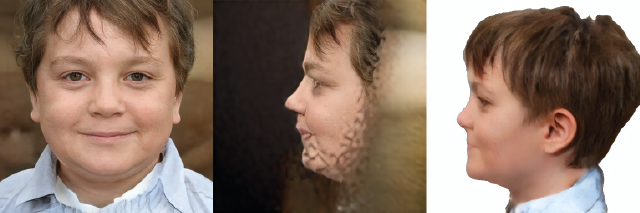}
     \makebox[0.3\linewidth][c]{Input}\hfill
     \makebox[0.3\linewidth][c]{Eg3D~\cite{Chan_2022_CVPR}}\hfill          
     \makebox[0.3\linewidth][c]{Ours}\hfill
  \caption{\ourname{} and Eg3D~\cite{Chan_2022_CVPR} are shown at $+90^{\circ}$ angle.}
  \label{fig:eg3d}   
  \vspace{-0.5cm}
\end{figure}

\section{Qualitative comparison with Panohead and Eg3D}

Panohead~\cite{an2023panohead} is a NeRF-based method capable of generating $360^{\circ}$ views.
Given an input facial image, it requires a fitting process to produce novel views, often necessitating additional pivotal tuning.
In contrast, \ourname{} eliminates the need for any fitting or fine-tuning steps.
Additionally, as presented in Fig.~\ref{fig:panohead}, Panohead frequently introduces artifacts on the back of the head, a limitation our method overcomes.
On the other hand,  EG3D~\cite{Chan_2022_CVPR} is another NeRF-based method having similar drawbacks as Panohead.
Moreover, it only focuses on generating near-frontal views contrary to our full-head approach as shown in Fig~\ref{fig:eg3d}.
We present a qualitative comparison between our method and Panohead~\cite{an2023panohead} in Fig.~\ref{fig:comparison_panohead}. As demonstrated, our proposed approach achieves more faithful novel view synthesis of the input subjects and significantly reduces the back-of-head artifacts commonly observed in Panohead.

\vspace{-0.5cm}

\begin{figure}[!t]
    \centering
    \includegraphics[width=\linewidth]{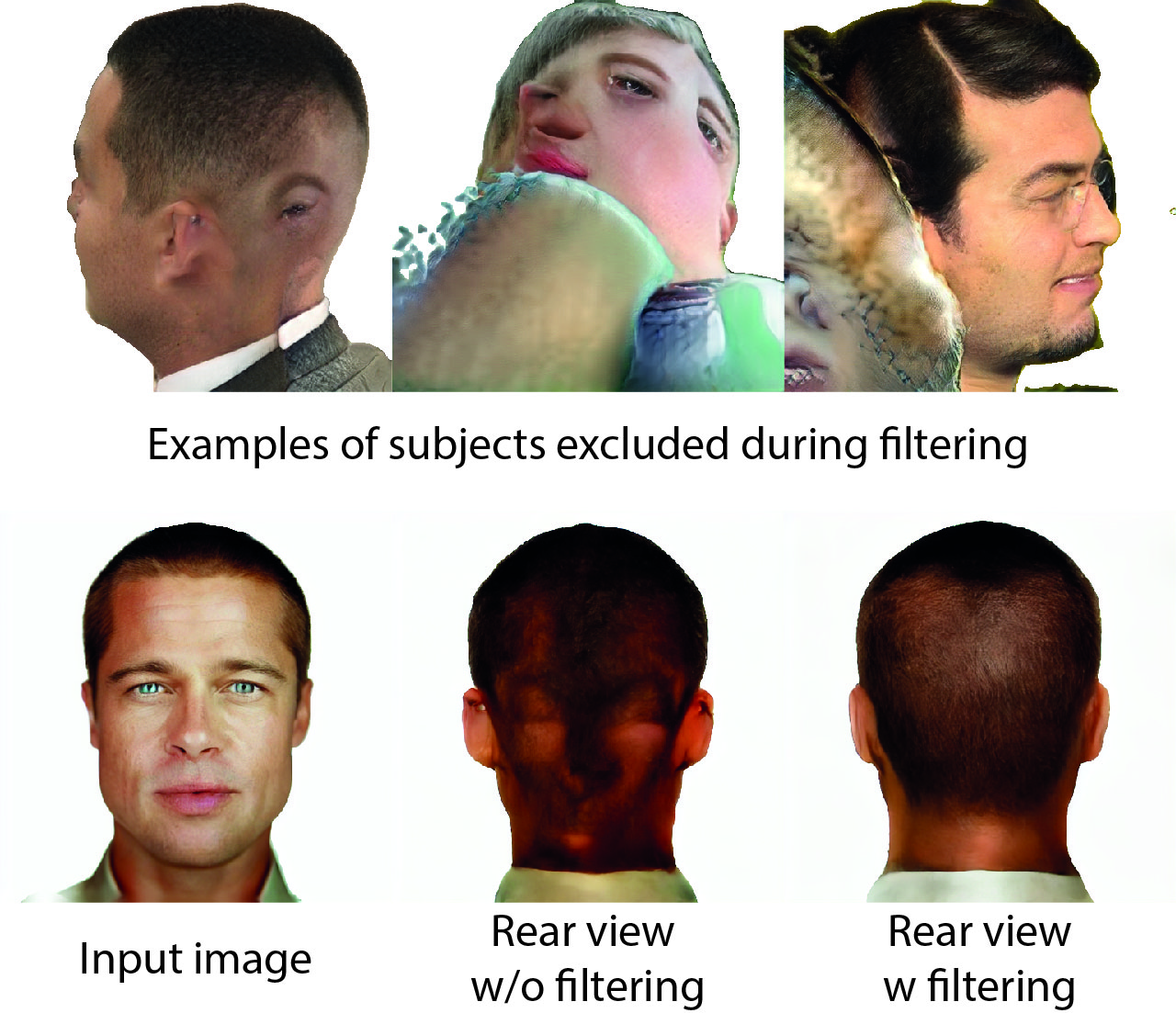}
    \caption{Top row: Examples of filtered-out subjects. 
    Bottom row: We present the rear view of an input image (left) when given to a model trained with the unfiltered and the filtered datasets.}
    \label{fig:ablation-dataset}
    \vspace{-0.5cm}
\end{figure}

\paragraph{Importance of the used dataset}

To train our method, we utilize a synthetic dataset, acquired using Panohead~\cite{an2023panohead}. 
Since artifacts frequently occur during sampling, a manual filtering step is necessary.
We filtered out failure cases, including artifacts or abnormal side-heads, that could negatively impact the performance of our model. We present excluded samples in the first row of Figure~\ref{fig:ablation-dataset}.
To confirm this, we trained two models for 100 epochs: (a) one using the prior-filtering dataset, and (b) one using the filtered dataset. 
Then, we gathered 50 images and compared their identity similarity scores at a yaw angle of $45^{\circ}$. 
The first model achieved an identity similarity score of $0.651$, whilst the model trained on the filtered dataset achieved a score of $0.72$.
Moreover, we showcase the generated real views of a representative subject in the bottom row of Figure~\ref{fig:ablation-dataset}.
\section{Identity sampling}

As mentioned in Section 5.1 and presented in Figure~8 of the main paper, our method can generate multi-view human identities, given only the input embedding. 
Although, this work does not focus on multi-view identity sampling, we explore our method's capabilities in this section.

As \ourname{} has been trained using the classifier-free guidance (CFG)\cite{ho2022classifier} whilst getting 0, 1 or 3 conditioning input images, it can be used to conditionally generate novel images depicting a similar identity as the input one.
By setting the guidance scale equal with 3.5, we run the EDM sampler~\cite{karras2022edm} for 50 sampling steps.
We set $S_{churn}=0, S_{tmin}=0.05, S_{tmax}=50, S_{noise}=1.003$ and we use the EDM~\cite{karras2022edm} discretization steps, with maximum sigma equal to 700.
The sampling process takes about 10 seconds while it runs on an NVIDIA A100-PCIE.
We present samples generated from our model in Figure~\ref{fig:cfgsampling}.

\section{More samples}

We provide additional results in Figures~\ref{fig:extreme_angles}, ~\ref{fig:comparison_nersemble_all}, ~\ref{fig:elevation}, ~\ref{fig:samples} and \ref{fig:samples2}. 
\ourname{} is demonstrated under extreme viewpoints in Figure~\ref{fig:extreme_angles} whilst Figure~\ref{fig:comparison_nersemble_all} presents a qualitative comparison between our method's performance and Panohead~\cite{an2023panohead}, Eg3D~\cite{Chan_2022_CVPR}, DiffPortrait3D~\cite{Gu_2024_CVPR} and SV3D~\cite{voleti2024sv3dnovelmultiviewsynthesis}.
In Figure~\ref{fig:elevation}, we showcase samples generated while using \ourname{}, under $\{ \pm 9^{\circ}, \pm 16^{\circ}, \pm 23^{\circ} \}$ elevation and azimuth angles.
Additionally, samples produced from our model are presented in Figures ~\ref{fig:samples} and \ref{fig:samples2}, given the input images on the left. 
As illustrated, our proposed methodology can be applied to a wide variety of images, including diverse identities, input angles and image styles.

\begin{figure*}[!t]
    \centering
    \includegraphics[width=\linewidth]{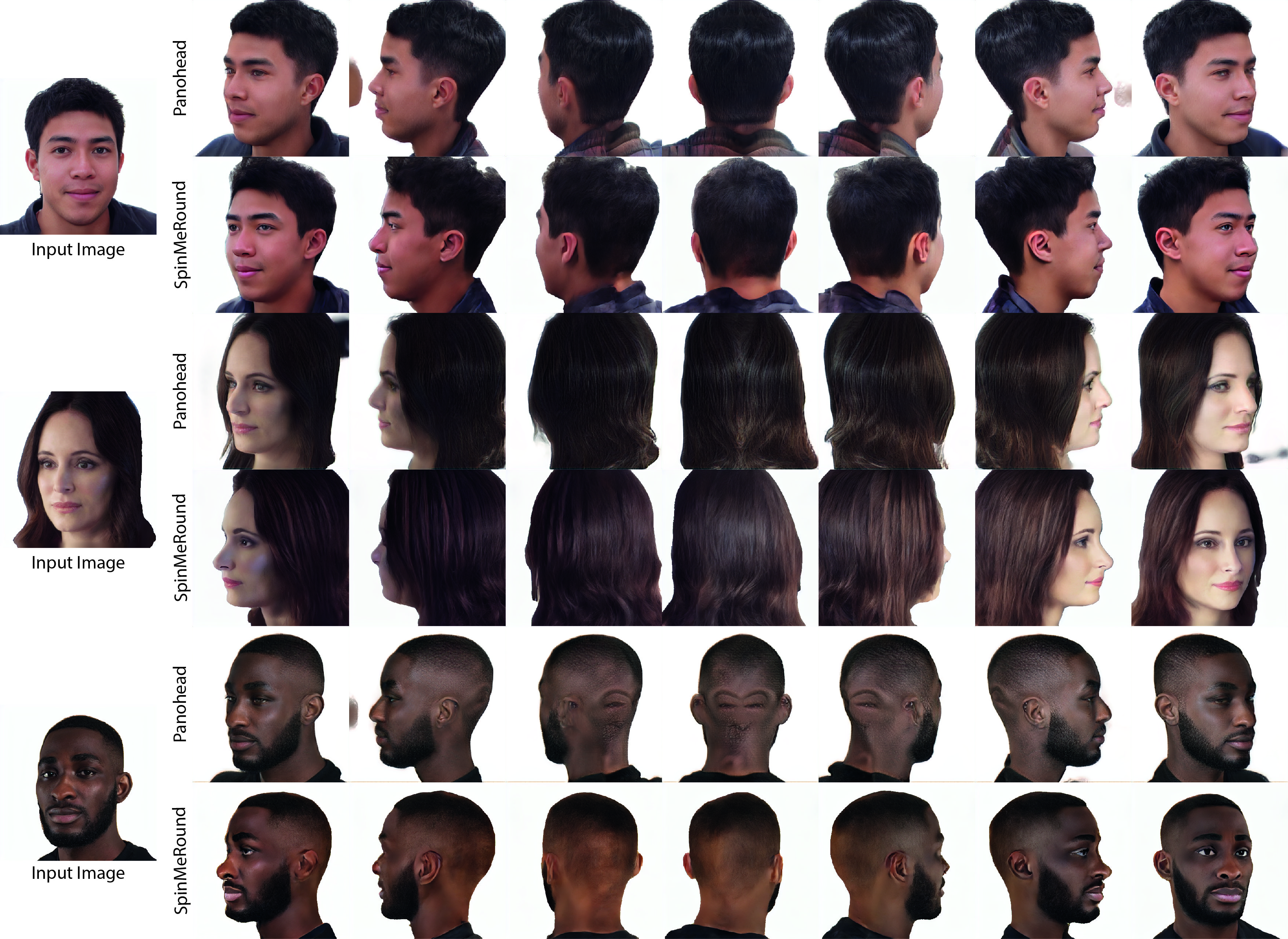}
    \begin{scriptsize}
        \hspace*{2.75cm}\makebox[0.12\linewidth][c]{$+45^{\circ}$}\hfill
        \makebox[0.12\linewidth][c]{$+90^{\circ}$}\hfill
        \makebox[0.12\linewidth][c]{$+135^{\circ}$}\hfil
        \makebox[0.12\linewidth][c]{$+180^{\circ}$}\hfill
        \makebox[0.12\linewidth][c]{$+225^{\circ}$}\hfill
        \makebox[0.12\linewidth][c]{$+270^{\circ}$}\hfill
        \makebox[0.12\linewidth][c]{$+315^{\circ}$}\hfill
    \end{scriptsize}
    \caption{We present a qualitative comparison between \ourname{} and Panohead~\cite{an2023panohead} at yaw angles from  $+45^{\circ}$ to $+315^{\circ}$. Contrary to Panodhead, our method faithfully generates novel views of the input subjects, without any artifacts on the backhead.}
    \label{fig:comparison_panohead}
\end{figure*}

\begin{figure*}[!t]
    \centering
    \includegraphics[width=\linewidth]{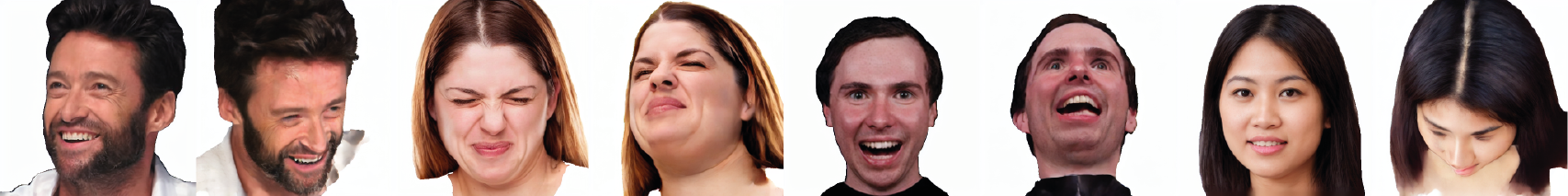}
    \begin{scriptsize}
        \makebox[0.12\linewidth][c]{Input image}\hfill
        \makebox[0.12\linewidth][c]{Novel view}\hfill
        \makebox[0.12\linewidth][c]{Input image}\hfil
        \makebox[0.12\linewidth][c]{Novel view}\hfill
        \makebox[0.12\linewidth][c]{Input image}\hfill
        \makebox[0.12\linewidth][c]{Novel view}\hfill
        \makebox[0.12\linewidth][c]{Input image}\hfill
        \makebox[0.12\linewidth][c]{Novel view}\hfill
    \end{scriptsize}
    \caption{\ourname{} is showcased under extreme angles.}
    \label{fig:extreme_angles}
\end{figure*}

\begin{figure*}[!t]
\begin{center}
    \includegraphics[width=\textwidth]{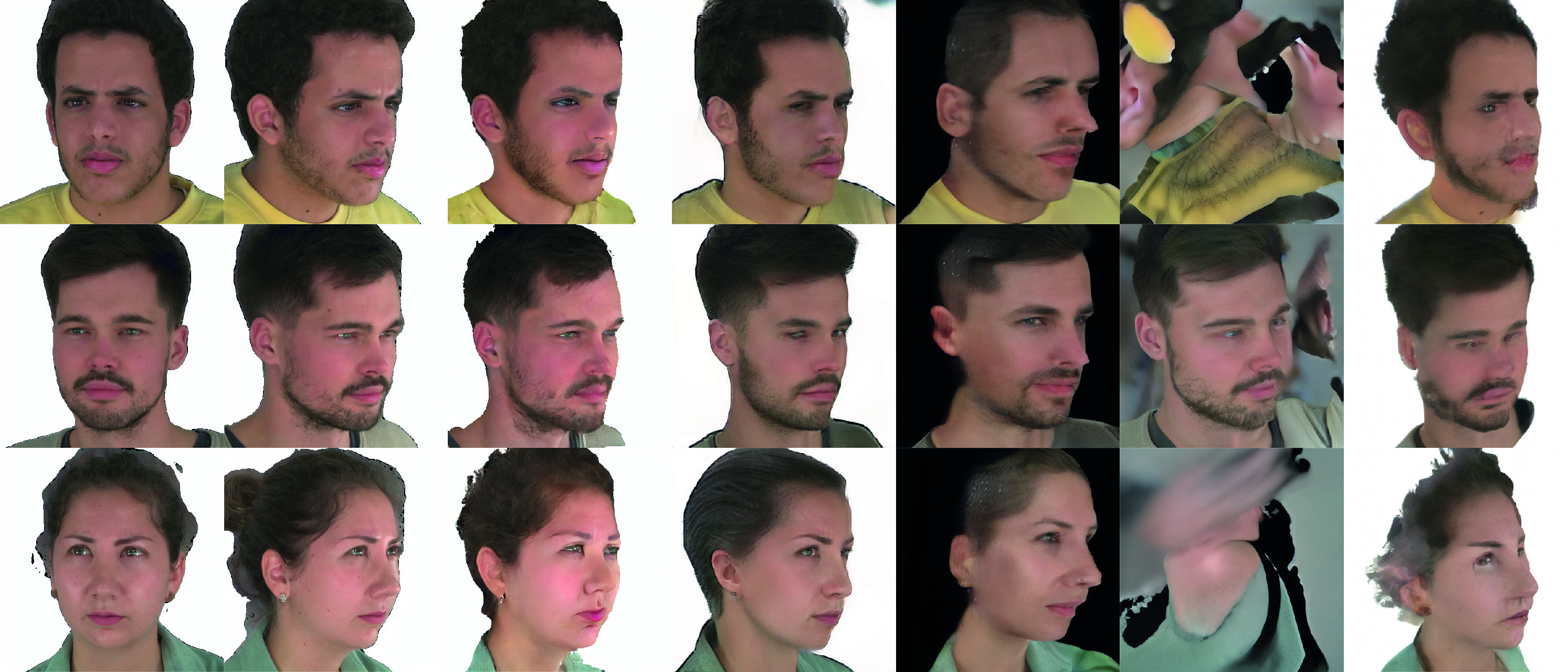}
    \begin{scriptsize}
    \makebox[0.14\linewidth][c]{Input image}\hfill
    \makebox[0.08\linewidth][c]{Ground-truth}\hfill          
    \makebox[0.16\linewidth][c]{\ourname{} (Ours)}\hfill
    \makebox[0.12\linewidth][c]{Panohead~\cite{an2023panohead}}\hfill
    \makebox[0.16\linewidth][c]{Eg3D~\cite{Chan_2022_CVPR}}\hfill
    \makebox[0.12\linewidth][c]{DiffPortrait3D~\cite{Gu_2024_CVPR}}\hfill
    \makebox[0.12\linewidth][c]{SV3D~\cite{voleti2024sv3dnovelmultiviewsynthesis}}\hfill
    \end{scriptsize}
    \caption{Qualitative comparison on NeRSemble dataset~\cite{kirschstein2023nersemble}. 
    As showcased, \ourname{} shows superior results over Panohead~\cite{an2023panohead}, Eg3D~\cite{Chan_2022_CVPR}, DiffPortrait3D~\cite{Gu_2024_CVPR} and SV3D~\cite{voleti2024sv3dnovelmultiviewsynthesis}.}
    \label{fig:comparison_nersemble_all}
\end{center}
\end{figure*}

\begin{figure*}[!t]
\begin{center}
    \includegraphics[width=\textwidth]{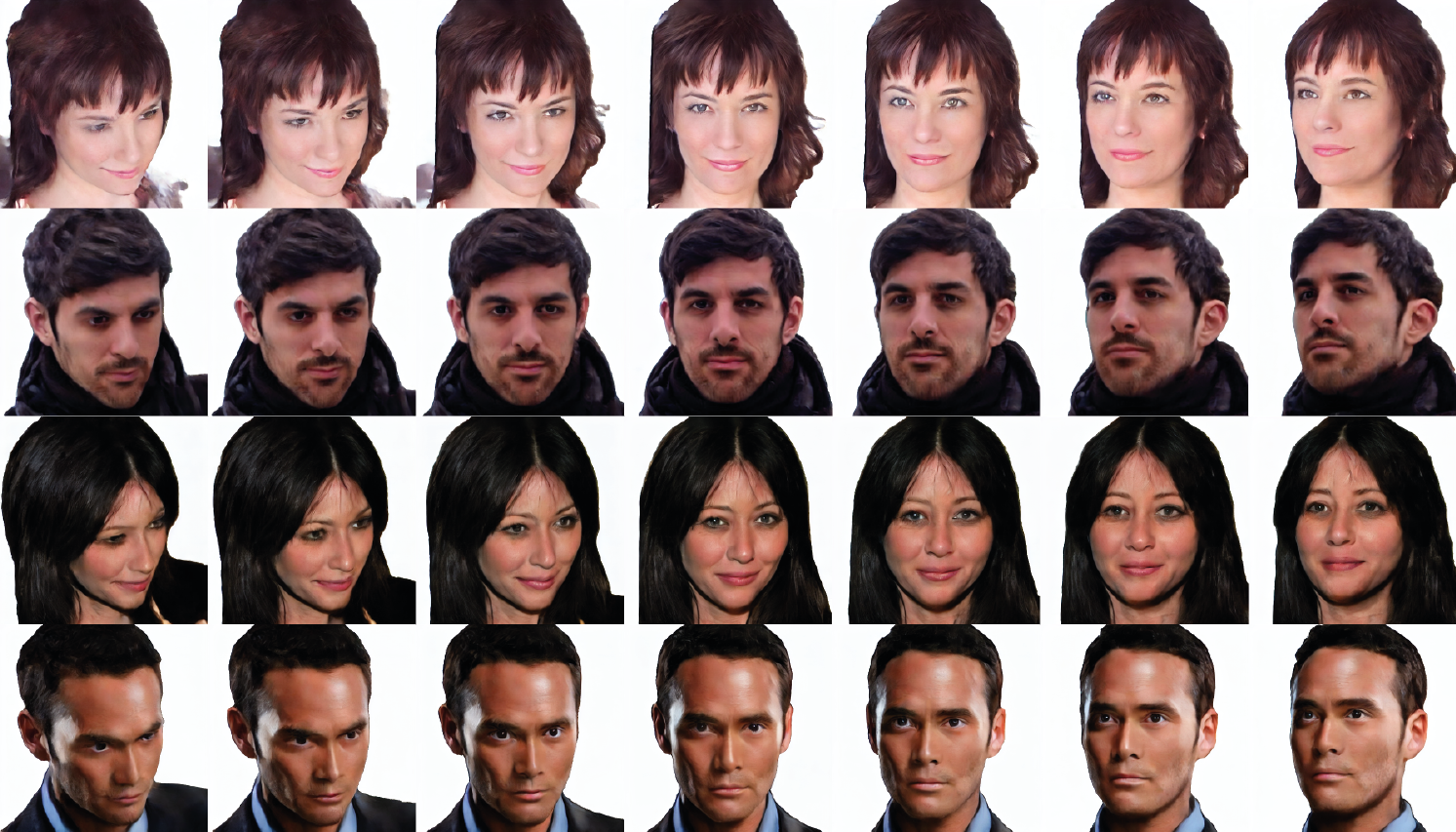}
    \makebox[0.12\linewidth][c]{$-23^{\circ}$}\hfill
     \makebox[0.12\linewidth][c]{$-16^{\circ}$}\hfill          
     \makebox[0.12\linewidth][c]{$-9^{\circ}$}\hfill
    \makebox[0.12\linewidth][c]{Input}\hfill
    \makebox[0.12\linewidth][c]{$+9^{\circ}$}\hfill
    \makebox[0.12\linewidth][c]{$+16^{\circ}$}\hfill
    \makebox[0.12\linewidth][c]{$+23^{\circ}$}\hfill
    \caption{We showcase samples under $\{ \pm 9^{\circ}, \pm 16^{\circ}, \pm 23^{\circ} \}$ elevation and azimuth angles. }
    \label{fig:elevation}
\end{center}
\end{figure*}

\def \var {1}
\begin{figure*}[!t]
\begin{center}

\begin{subfigure}[b]{\var\textwidth}
    \includegraphics[width=\textwidth]{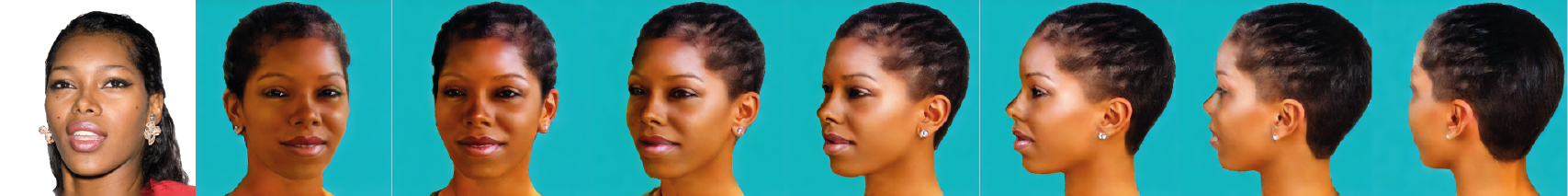}
\end{subfigure}
\begin{subfigure}[b]{\var\textwidth}
    \includegraphics[width=\textwidth]{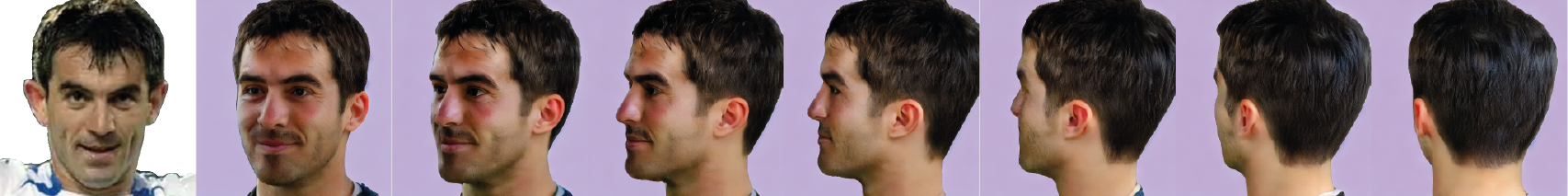}
\end{subfigure}
\begin{subfigure}[b]{\var\textwidth}
    \includegraphics[width=\textwidth]{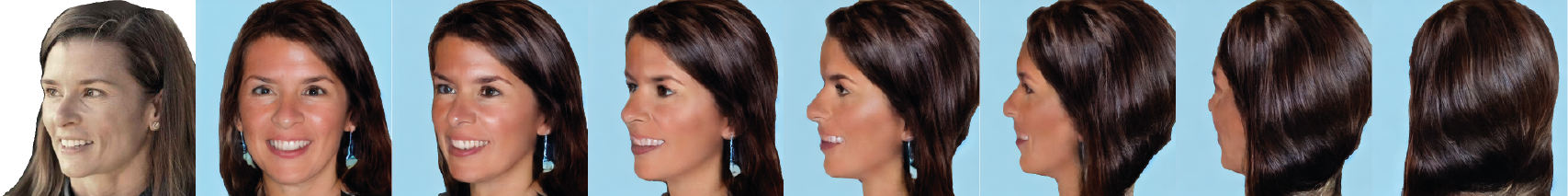}
\end{subfigure}
\begin{subfigure}[b]{\var\textwidth}
    \includegraphics[width=\textwidth]{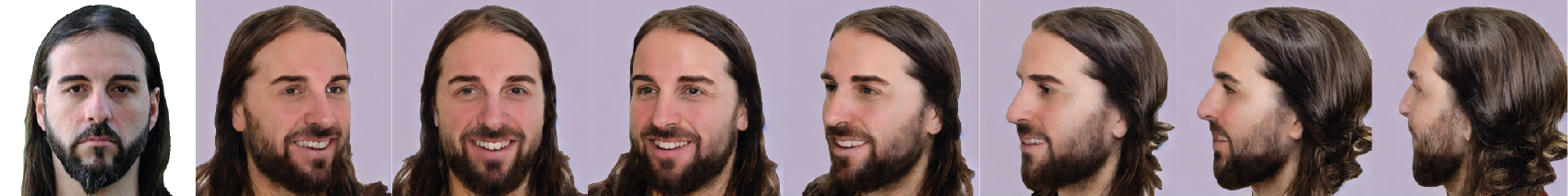}
\end{subfigure}
\begin{subfigure}[b]{\var\textwidth}
    \includegraphics[width=\textwidth]{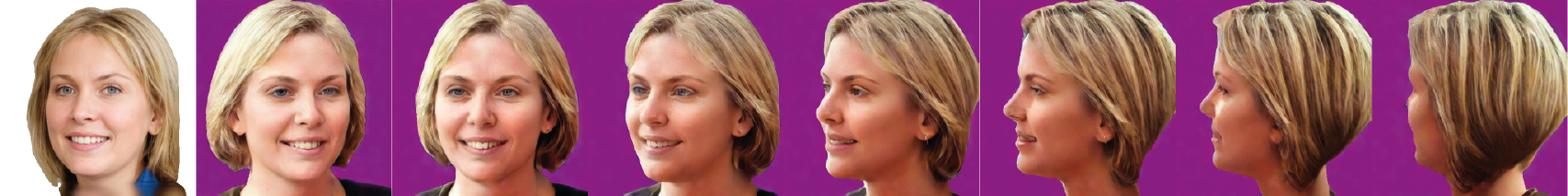}
\end{subfigure}
\begin{subfigure}[b]{\var\textwidth}
    \includegraphics[width=\textwidth]{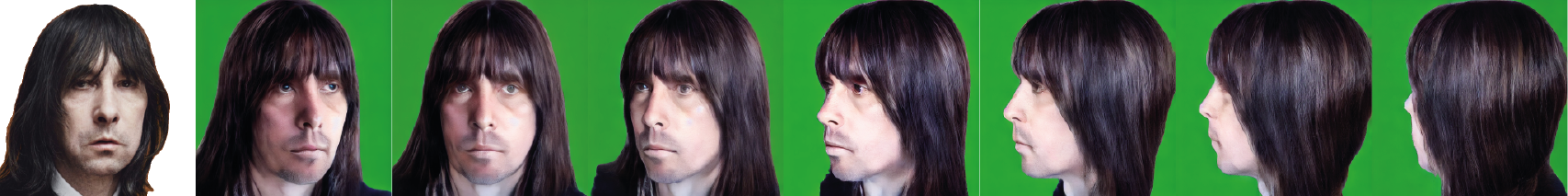}
\end{subfigure}
\begin{subfigure}[b]{\var\textwidth}
    \includegraphics[width=\textwidth]{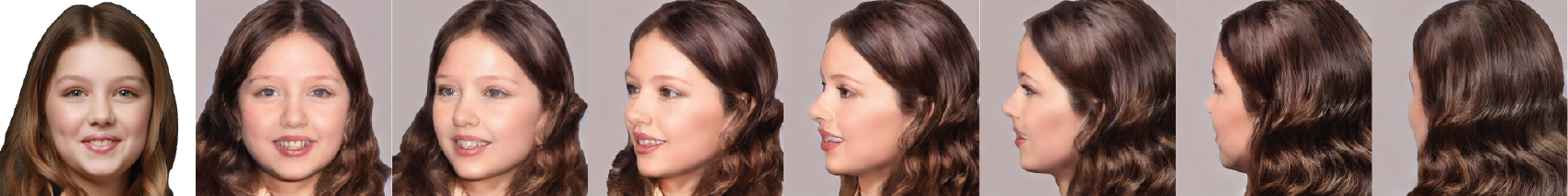}
\end{subfigure}
\begin{subfigure}[b]{\var\textwidth}
    \includegraphics[width=\textwidth]{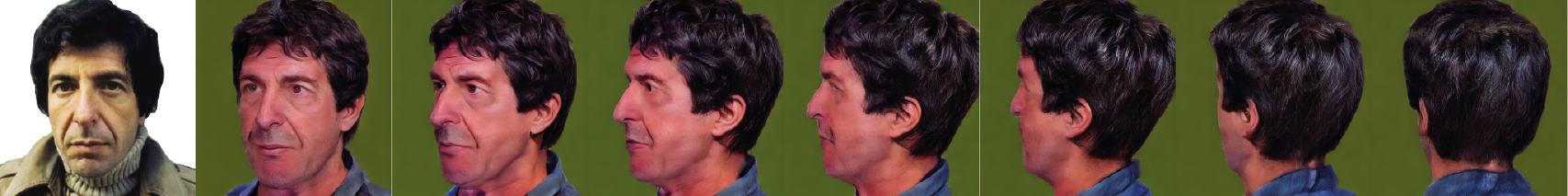}
\end{subfigure}
\begin{subfigure}[b]{\var\textwidth}
    \includegraphics[width=\textwidth]{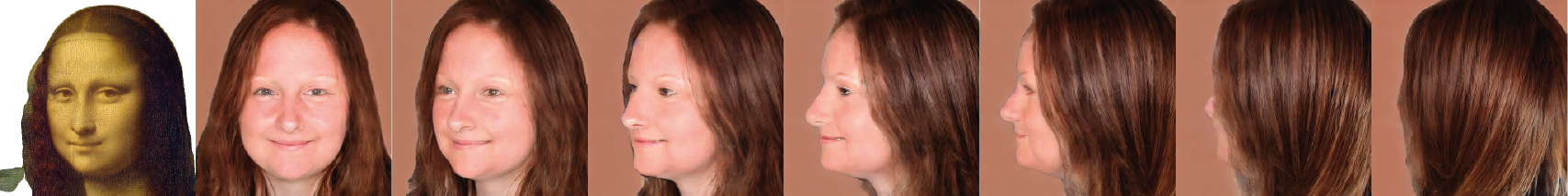}
\end{subfigure}
\begin{scriptsize}
     \makebox[0.15\linewidth][c]{Input Images}\hfill
     \makebox[0.70\linewidth][c]{Generated samples}\hfill
\end{scriptsize}
\caption{Samples generated using \ourname{} using \textit{only} the input identity vector.}
\label{fig:cfgsampling}
\end{center}
\end{figure*}

\def \var {1}
\begin{figure*}[!h]
\begin{center}
 \begin{subfigure}[b]{\var\textwidth}
      \includegraphics[width=\textwidth]{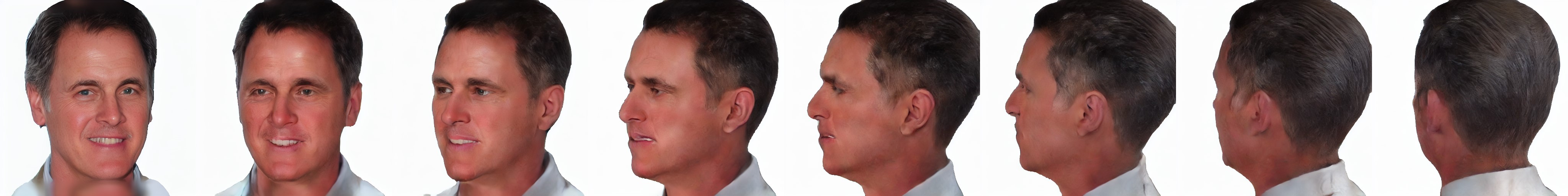}
 \end{subfigure}
  \begin{subfigure}[b]{\var\textwidth}
      \includegraphics[width=\textwidth]{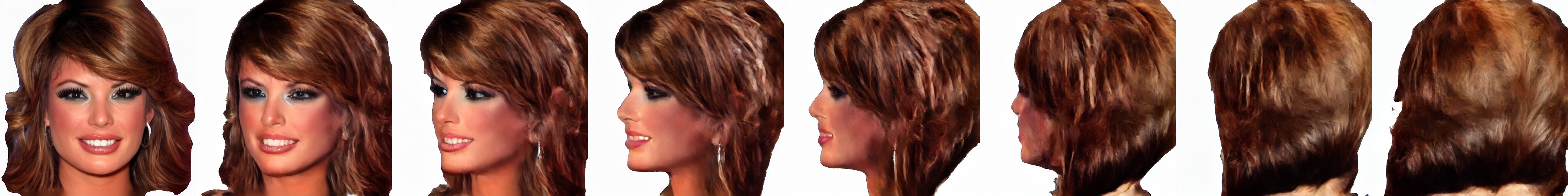}
 \end{subfigure}
  \begin{subfigure}[b]{\var\textwidth}
      \includegraphics[width=\textwidth]{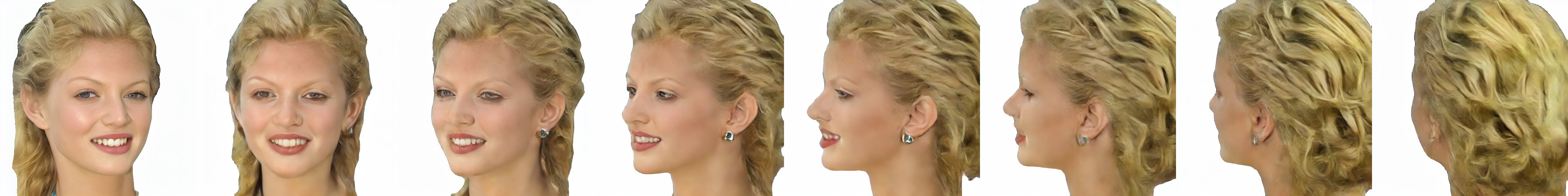}
 \end{subfigure}
  \begin{subfigure}[b]{\var\textwidth}
      \includegraphics[width=\textwidth]{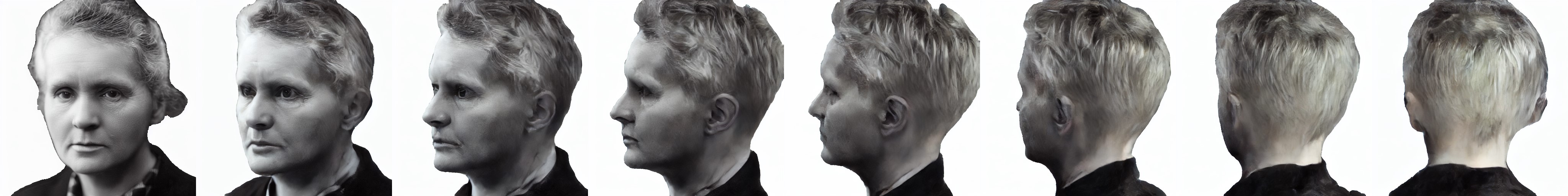}
 \end{subfigure}
  \begin{subfigure}[b]{\var\textwidth}
      \includegraphics[width=\textwidth]{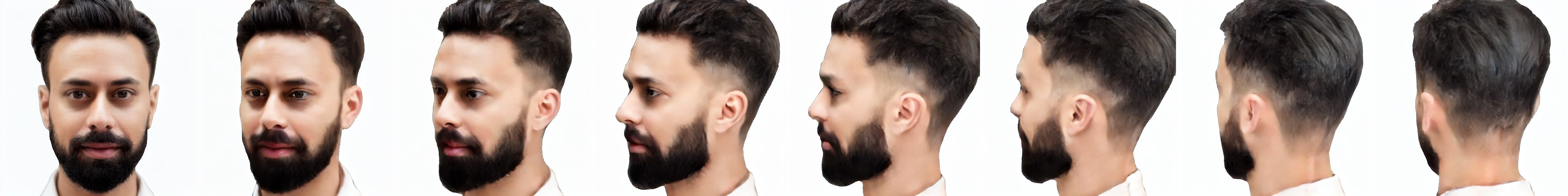}
 \end{subfigure}
  \begin{subfigure}[b]{\var\textwidth}
      \includegraphics[width=\textwidth]{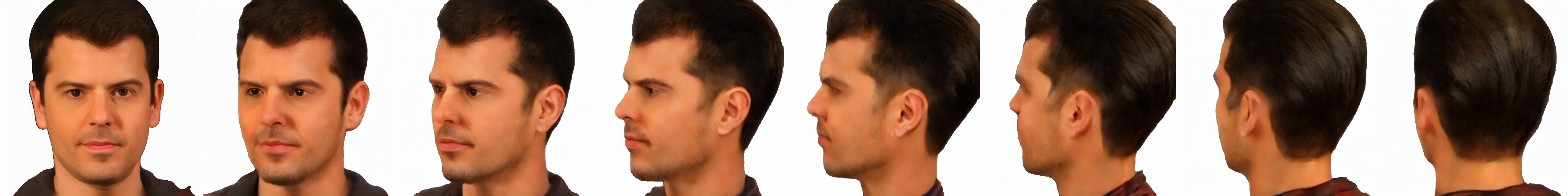}
 \end{subfigure}
  \begin{subfigure}[b]{\var\textwidth}
      \includegraphics[width=\textwidth]{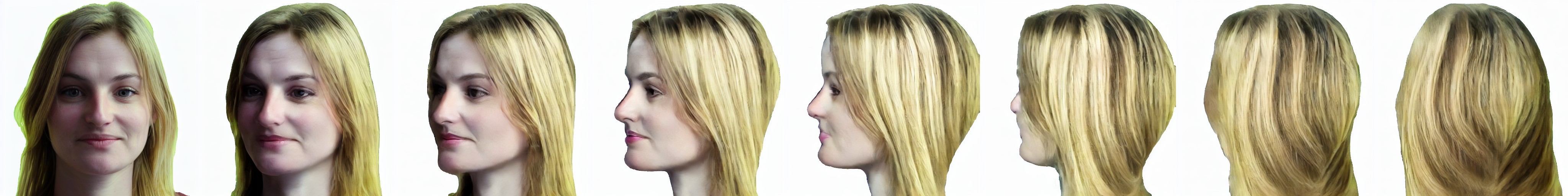}
 \end{subfigure}
  \begin{subfigure}[b]{\var\textwidth}
      \includegraphics[width=\textwidth]{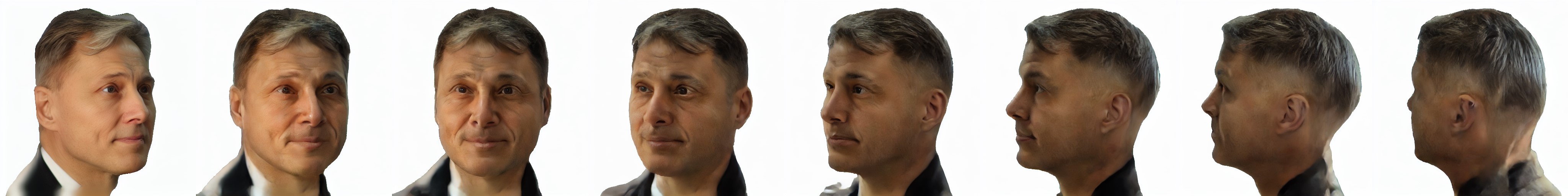}
 \end{subfigure}
  \begin{subfigure}[b]{\var\textwidth}
      \includegraphics[width=\textwidth]{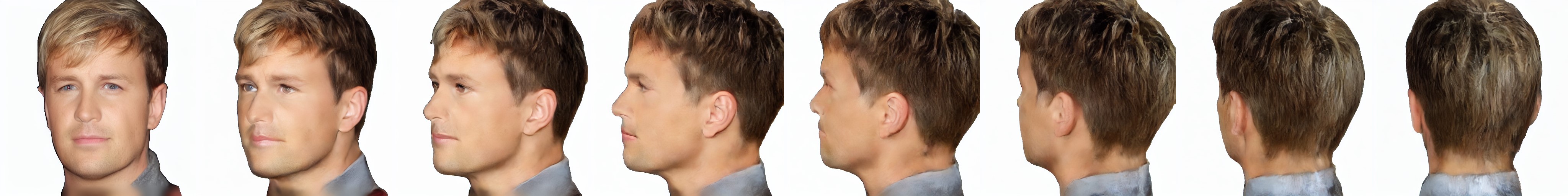}
 \end{subfigure}
\begin{scriptsize}
     \makebox[0.15\linewidth][c]{Input Images}\hfill
     \makebox[0.70\linewidth][c]{Generated samples}\hfill
\end{scriptsize}
\caption{Samples generated with our method, using the images on the left as input (1/2).}
\label{fig:samples}
\end{center}
\end{figure*}

\begin{figure*}[!h]
\begin{center}
  \begin{subfigure}[b]{\var\textwidth}
      \includegraphics[width=\textwidth]{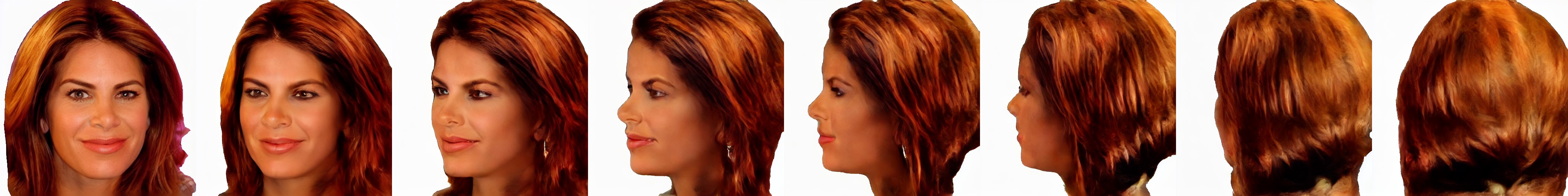}
 \end{subfigure}
  \begin{subfigure}[b]{\var\textwidth}
      \includegraphics[width=\textwidth]{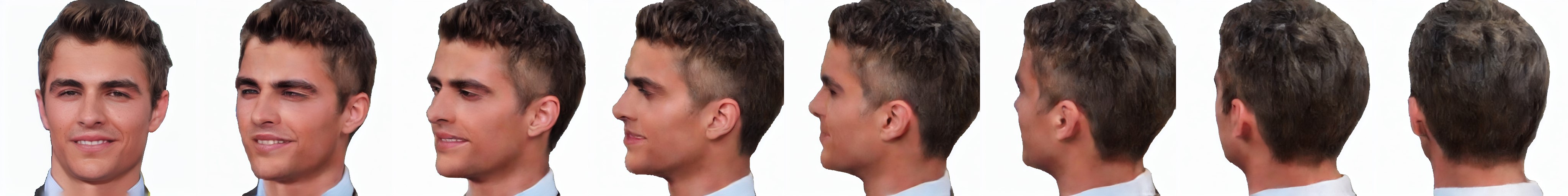}
 \end{subfigure}
  \begin{subfigure}[b]{\var\textwidth}
      \includegraphics[width=\textwidth]{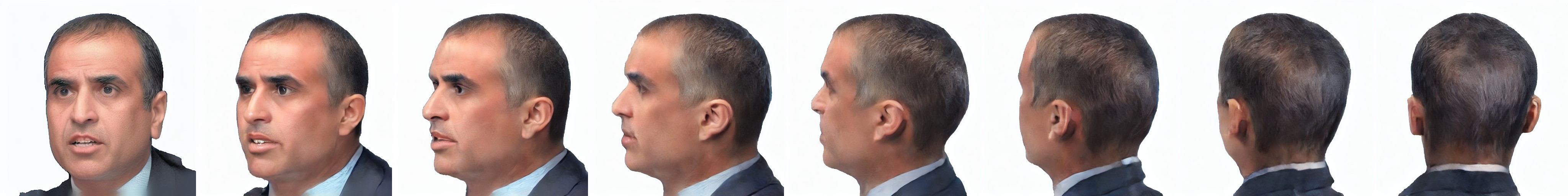}
 \end{subfigure}
  \begin{subfigure}[b]{\var\textwidth}
      \includegraphics[width=\textwidth]{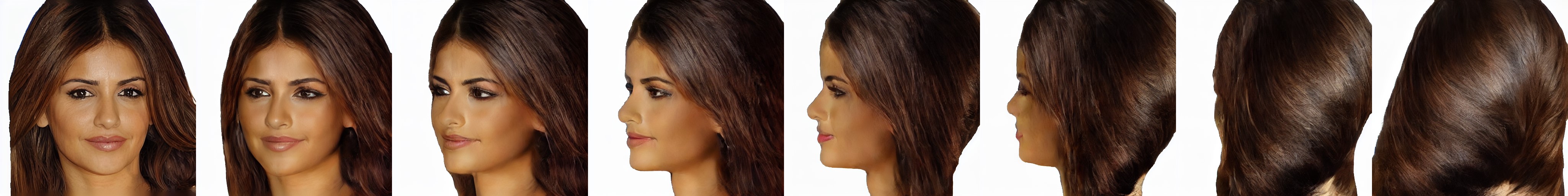}
 \end{subfigure}
  \begin{subfigure}[b]{\var\textwidth}
      \includegraphics[width=\textwidth]{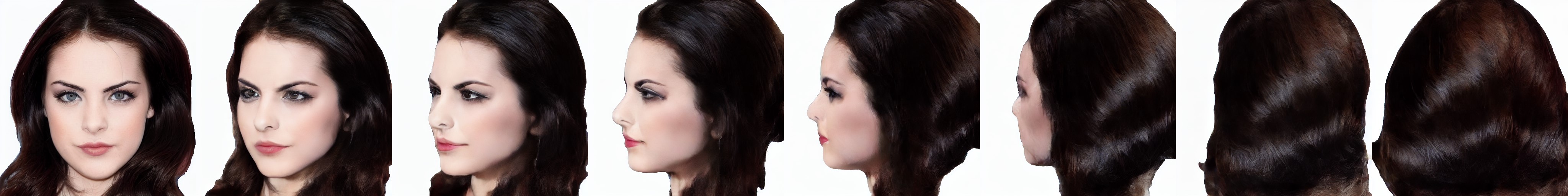}
 \end{subfigure}
  \begin{subfigure}[b]{\var\textwidth}
      \includegraphics[width=\textwidth]{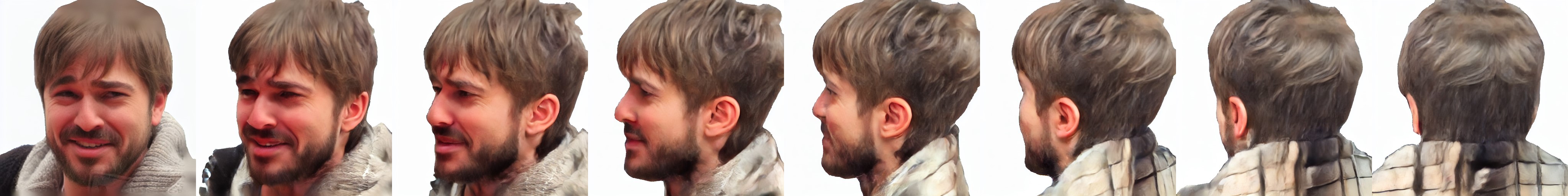}
 \end{subfigure}
  \begin{subfigure}[b]{\var\textwidth}
      \includegraphics[width=\textwidth]{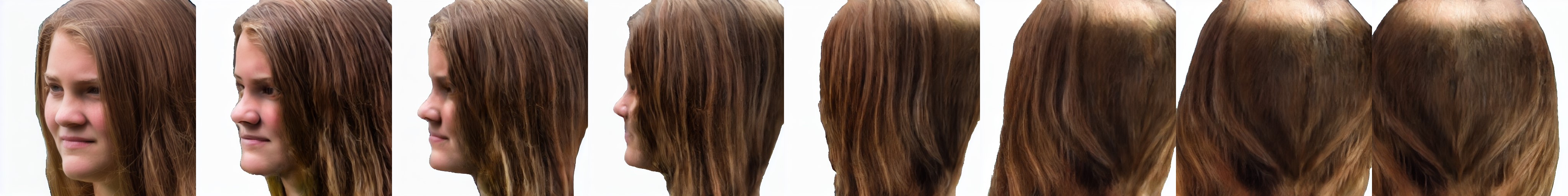}
 \end{subfigure}
  \begin{subfigure}[b]{\var\textwidth}
      \includegraphics[width=\textwidth]{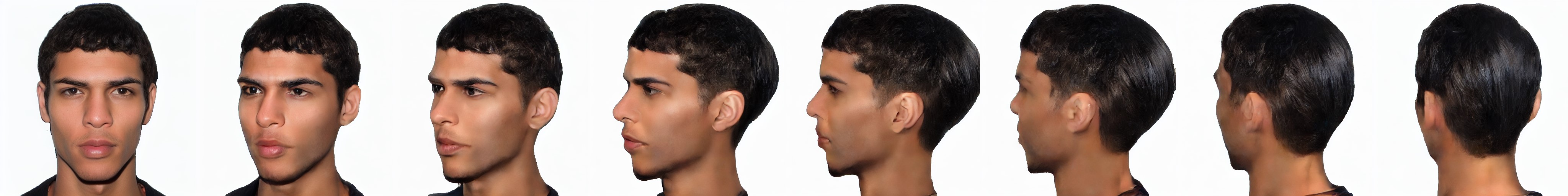}
 \end{subfigure}
  \begin{subfigure}[b]{\var\textwidth}
      \includegraphics[width=\textwidth]{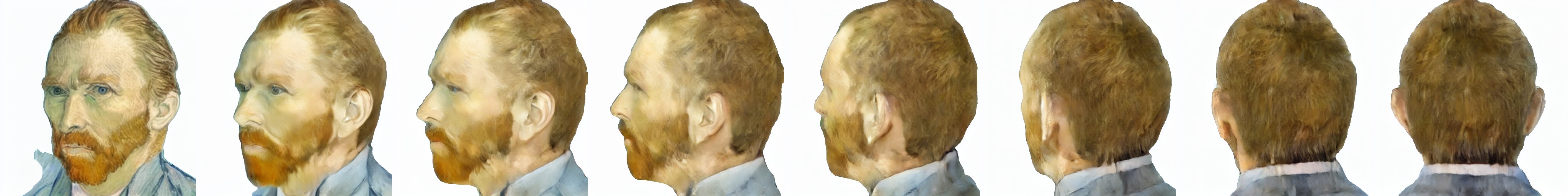}
 \end{subfigure}
\begin{scriptsize}
     \makebox[0.15\linewidth][c]{Input Images}\hfill
     \makebox[0.70\linewidth][c]{Generated samples}\hfill
\end{scriptsize}
\caption{Samples generated with our method, using the images on the left as input (2/2).}
\label{fig:samples2}
\end{center}
\end{figure*}

\end{document}